\documentclass[journal]{IEEEtai}
\usepackage{amsmath,amsfonts}
\usepackage{algorithmic}
\usepackage{algorithm}
\usepackage{array}
\usepackage[caption=false,font=normalsize,labelfont=sf,textfont=sf]{subfig}
\usepackage{textcomp}
\usepackage{stfloats}
\usepackage{url}
\usepackage{verbatim}
\usepackage{graphicx}
\usepackage{cite}
\usepackage{bm}
\usepackage[table]{xcolor}
\usepackage{xcolor}
\usepackage{array}
\usepackage{tabularx}
\usepackage{fancyhdr}
\usepackage{algorithm}
\usepackage{algorithmic}
\usepackage{multirow}
\usepackage{booktabs}
\usepackage{array} 
\usepackage{amsthm}
\usepackage{bm}
\usepackage{amssymb}
\usepackage{bbding}
\usepackage{pifont}
\usepackage{wasysym}
\usepackage{amssymb}
\usepackage{graphicx}
\usepackage{makecell}
\usepackage{color}
\usepackage{caption}
\usepackage{xcolor}
\usepackage{subcaption}
\usepackage{multicol}
\usepackage{hyperref}

\hypersetup{hidelinks,
	colorlinks=true,
	allcolors=black,
	pdfstartview=Fit,
	breaklinks=true}
\begin{document}
\title{CoT-Drive: Efficient Motion Forecasting for Autonomous Driving with LLMs and Chain-of-Thought Prompting}

\markboth{JOURNAL OF XXXX}
{Author \MakeLowercase{\textit{et al.}}: Large Language Model for Motion Forecasting}

\author{Haicheng~Liao$^{*}$,
        Hanlin~Kong$^{*}$,
        Bonan~Wang,
        Chengyue~Wang,
        Wang~Ye,\\
        Zhengbing~He, ~\IEEEmembership{Senior Member, ~IEEE},
        Chengzhong~Xu, ~\IEEEmembership{Fellow, ~IEEE}
        and Zhenning Li$^{\dag}$
\thanks{\dag\,Corresponding author; *\,Authors contributed equally.} \thanks{Haicheng Liao, Hanlin Kong, Bonan Wang, Chengyue Wang, Chengzhong Xu, and Zhenning Li are with the State Key Laboratory of Internet of Things for Smart City, University of Macau, Macau. Ye Wang is with the Department of Computer and Information Science, University of Macau, Macau. Zhengbing He is with Senseable City Lab, Massachusetts Institute of Technology, Cambridge MA, United States. E-mails: zhenningli@um.edu.mo.}

\thanks{This research is supported by the  State Key Lab of Intelligent Transportation System under Project (2024-B001), Science and Technology Development Fund of Macau SAR (File no. 0021/2022/ITP, 0081/2022/A2, 001/2024/SKL), Shenzhen-Hong Kong-Macau Science and Technology Program Category C (SGDX20230821095159012), and University of Macau (SRG2023-00037-IOTSC).}}

\maketitle
{ 
\begin{abstract}
Accurate motion forecasting is crucial for safe autonomous driving (AD). This study proposes CoT-Drive, a novel approach that enhances motion forecasting by leveraging large language models (LLMs) and a chain-of-thought (CoT) prompting method.  We introduce a teacher-student knowledge distillation strategy to effectively transfer LLMs' advanced scene understanding capabilities to lightweight language models (LMs), ensuring that CoT-Drive operates in real-time on edge devices while maintaining comprehensive scene understanding and generalization capabilities. By leveraging CoT prompting techniques for LLMs without additional training, CoT-Drive generates semantic annotations that significantly improve the understanding of complex traffic environments, thereby boosting the accuracy and robustness of predictions.  Additionally, we present two new scene description datasets, Highway-Text and Urban-Text, designed for fine-tuning lightweight LMs to generate context-specific semantic annotations.  Comprehensive evaluations of five real-world datasets demonstrate that CoT-Drive outperforms existing models, highlighting its effectiveness and efficiency in handling complex traffic scenarios. Overall, this study is the first to consider the practical application of LLMs in this field. It pioneers the training and use of a lightweight LLM surrogate for motion forecasting, setting a new benchmark and showcasing the potential of integrating LLMs into AD systems.

\end{abstract}}


\section{Introduction} \label{Introduction}
{{
\begin{figure}[t]
\centering
\includegraphics[width=0.49\textwidth]{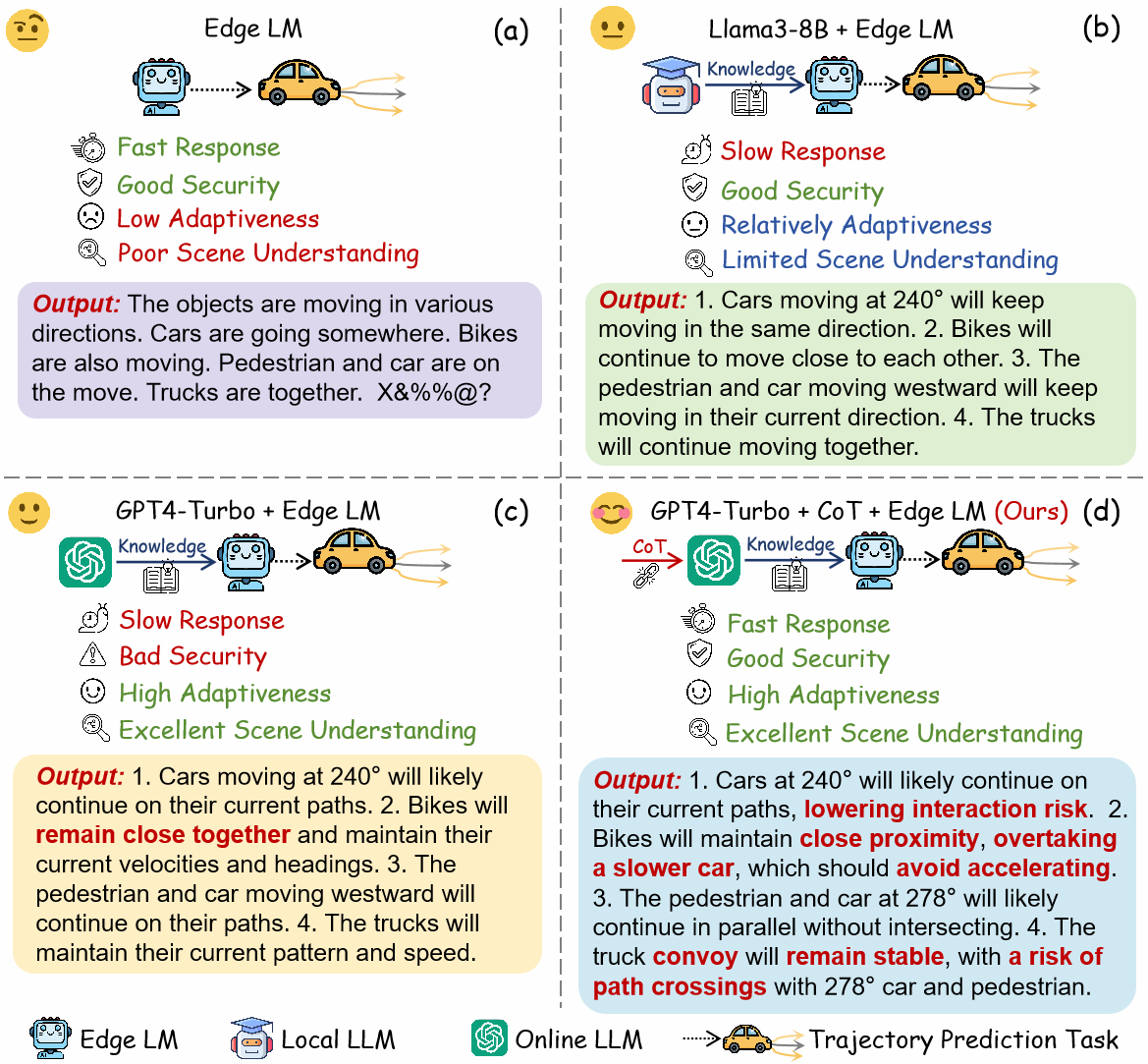}
\caption{{Illustration of the strength of COT-Drive (d), comparing edge LMs (a), local LLMs with edge LMs (b), and online LLMs with edge LMs (c) on key perspectives: response time, security, adaptability, and scene understanding capabilities.}}
\label{fig:structure1}
\end{figure}

Accurate motion forecasting of traffic agents in dynamic and heterogeneous environments is fundamental for autonomous vehicles' (AVs) decision-making and safe planning, serving as a cornerstone for autonomous driving (AD) systems. \cite{liao2024real,li2024efficient} }These environments necessitate motion forecasting models that can effectively comprehend contextual semantic information, including the movements and behaviors of various agents—such as vehicles, pedestrians, and cyclists—as well as environmental factors such as traffic signals and road conditions, along with the intricate interactions among these entities. Despite recent leaps forward in deep learning models for motion forecasting \cite{liao2024bat,chen2022intention}, they often falter in unseen or corner-case scenarios, revealing critical limitations in generalizability and contextual understanding. These data-driven models typically focus on more common and simplistic scenarios and tend to perform poorly when faced with real-world environments that differ significantly from their training data. This raises an urgent question: \textit{how can we enhance the adaptability and robustness of these models?}

The rapid developments in large language models (LLMs) like GPT-4 \cite{nazir2023comprehensive} and Llama-2 \cite{touvron2023llama} have offered a new perspective for understanding complex traffic scenarios and sparked interest in their application for motion forecasting tasks \cite{chen2024driving}.  Recent studies \cite{liao2024gpt,wen2024dilu} have demonstrated that LLMs not only improve the performance of motion forecasting models in common traffic scenarios but also excel in complex situations that require a deep understanding of contextual subtleties and intricate interactions between traffic agents. These powerful generalization and reasoning capabilities position LLMs as a promising solution to the challenges faced in this field \cite{chen2024driving}. However, deploying LLMs in AVs poses significant practical obstacles. On the one hand, online LLMs like GPT-4 Turbo and Palm \cite{chowdhery2023palm} can provide comprehensive scene understanding while alleviating computational burdens on edge devices. Yet, these models are restricted by communication conditions, which can hinder their effectiveness in real-time decision-making, particularly in rural or underdeveloped areas with unstable network connections or high latency. This can lead to delayed responses that compromise passenger safety.

Moreover, passengers face substantial costs to access these online services,  such as API usage fees and subscription charges for premium features, creating further barriers to their widespread adoption in AVs.
Importantly, they are vulnerable to data leakage or malicious tampering during data transmission, which could seriously damage the safety and property of passengers. On the other hand, while offline LLMs such as Vicuna \cite{zheng2023judging} and Flan-T5 \cite{chung2024scaling} can effectively mitigate the risks of data privacy and transmission delays, they typically struggle to capture the dynamic and uncertain nature of complex traffic scenarios as skillfully and flexibly as online LLMs.
In addition, the high storage and computational requirements for locally deploying these models pose a significant challenge for resource-constrained edge devices in real-world AVs.

These multifaceted challenges motivate us to explore critical questions about the future of motion forecasting for AVs: \textit{How can we develop a model that operates in real-time on edge devices while maintaining comprehensive scene understanding and generalization capabilities?} To address these challenges, this study presents CoT-Drive, a novel framework designed to integrate the advanced scene understanding capabilities of LLMs into a lightweight, edge-deployable model. Figure \ref{fig:structure1} illustrates our proposed approach, showcasing how the Chain-of-Thought (CoT) prompting technique enhances contextual semantic analysis while guiding language models (LMs) to emulate LLMs in traffic scene comprehension. We introduce a teacher-student knowledge distillation strategy to transfer knowledge from powerful LLMs to lightweight LMs. Specifically, the LLM GPT-4 Turbo acts as a ``teacher'', imparting its advanced scene understanding capacity to the lightweight ``student'' model.  This ``student’’ model, i.e., LMs, integrated into our motion forecasting framework, enhances scene interpretation and generalization while minimizing computational and storage overheads associated with the direct use of local LLMs. To fully enhance the model’s scene understanding capabilities, we refine the ``teacher'' model LLM using CoT prompting techniques that align its insights with human-like cognitive processes in driving contexts, thereby enabling efficient and accurate predictions for AVs.

Overall, this study addresses several key research questions:
\begin{itemize}

\item \textbf{Q1:} How can CoT-Drive provide efficient and accurate motion forecasting in challenging scenes, such as highways, dense urban areas, and complex intersections?

\item \textbf{Q2:} How can knowledge distillation effectively transfer the advanced scene understanding ability of LLMs to lightweight models, ensuring efficient, high-accuracy predictions on computationally constrained edge devices?

\item \textbf{Q3:} Can CoT prompting be utilized to enhance the contextual understanding of LLMs in complex scenarios, thereby improving motion forecasting accuracy and reliability without additional fine-tuning?

\end{itemize}
This paper embarks on an innovative journey to address these research questions. We will explore and resolve these research questions (\textbf{A1-A3}) throughout the study.The paper is structured as follows: Section \ref{Related Work} reviews relevant literature, while Section \ref{Dataset} introduces the proposed datasets. Section \ref{Methodology} formulates the problem and presents CoT-Drive. Section \ref{Experiment} evaluates its performance in real-world datasets and examines research questions. Section \ref{Discussion} outlines limitations and future research directions, and Section \ref{Conclusion} summarizes the findings.

\section{Related Work} \label{Related Work}
{ 
\textbf{Motion Forecasting in Autonomous Driving.}
Recent advancements in motion forecasting for AVs have leveraged deep learning models to capture spatio-temporal interactions among traffic agents. Early works, such as the CS-LSTM \cite{deo2018convolutional}, emphasized the role of social interactions, marking a pivotal point in motion forecasting. Subsequent contributions like MFTraj \cite{ijcai2024p657} and NEST \cite{wang2024nest} refined trajectory accuracy through attention mechanisms and goal-centered approaches. Further innovations, such as GAN-based models for long-term prediction \cite{ijcai2024p756,liao2024cognitive} and wave superposition techniques in WSiP \cite{Wang_Wang_Yan_Wang_2023}, have also improved prediction precision. It is well recognized that the motion of traffic agents is strongly influenced by surrounding agents, and several recent works have addressed prediction accuracy by incorporating uncertainty in agent behavior \cite{liao2025minds,ijcai2024p811}. Moreover, recent studies in Reinforcement Learning (RL) and Evolutionary Learning (EL) \cite{sun2018probabilistic} apply techniques like Markov Decision Processes and evolutionary algorithms for motion forecasting and optimal policy learning. However, they often lack nuanced understanding and human-like reasoning, limiting their real-world adaptability. To address this, recent research has shifted from purely architectural innovations to simulating real-world driving behaviors and intentions. For instance, HLTP \cite{liao2024less} and WAKE \cite{wang2025wake} introduce cognitive strategy and wavelet transform theory to enhance the understanding of driving behavior. Inspired by human drivers’ dynamic allocation of visual attention, HLTP++ \cite{10468619} further incorporates dynamic pooling mechanisms to mimic human perception during driving. 

\textbf{Large Language Models in Autonomous Driving.} LLMs have shown remarkable promise in the field of AD, facilitating complex traffic scene analysis, human-like reasoning, and pragmatic decision-making \cite{10522953}. Integrating LLMs into AD marks a major advancement towards enhancing system interpretability and adaptability. Early models like DiLu \cite{wen2024dilu} and CAVG \cite{liao2024gpt} pioneer the use of LLMs to process vast driving datasets, yet often struggled with generalizing across diverse environments. Recent advancements, such as Traj-LLM \cite{lan2024traj}, have improved multimodal understanding and context-aware reasoning. Notable methods extend this paradigm by integrating visual LLMs or using question-answer formats to mimic human-like driving, while Co-Pilot \cite{copilot} combines LLMs with feature engineering for better decision-making. Building on these foundations, studies like W3AL \cite{liao2024and} further refine scene understanding and reasoning. However, these LLM-based methods face challenges such as high computational costs, sensitivity to hyperparameters, and deployment difficulties in real-world autonomous driving. Online LLMs introduce latency and privacy concerns, limiting real-time feasibility. Overall, future work should balance model complexity and efficiency to optimize prediction accuracy for AD.

\begin{figure*}[tb]
\centering
\includegraphics[width=0.9\textwidth]{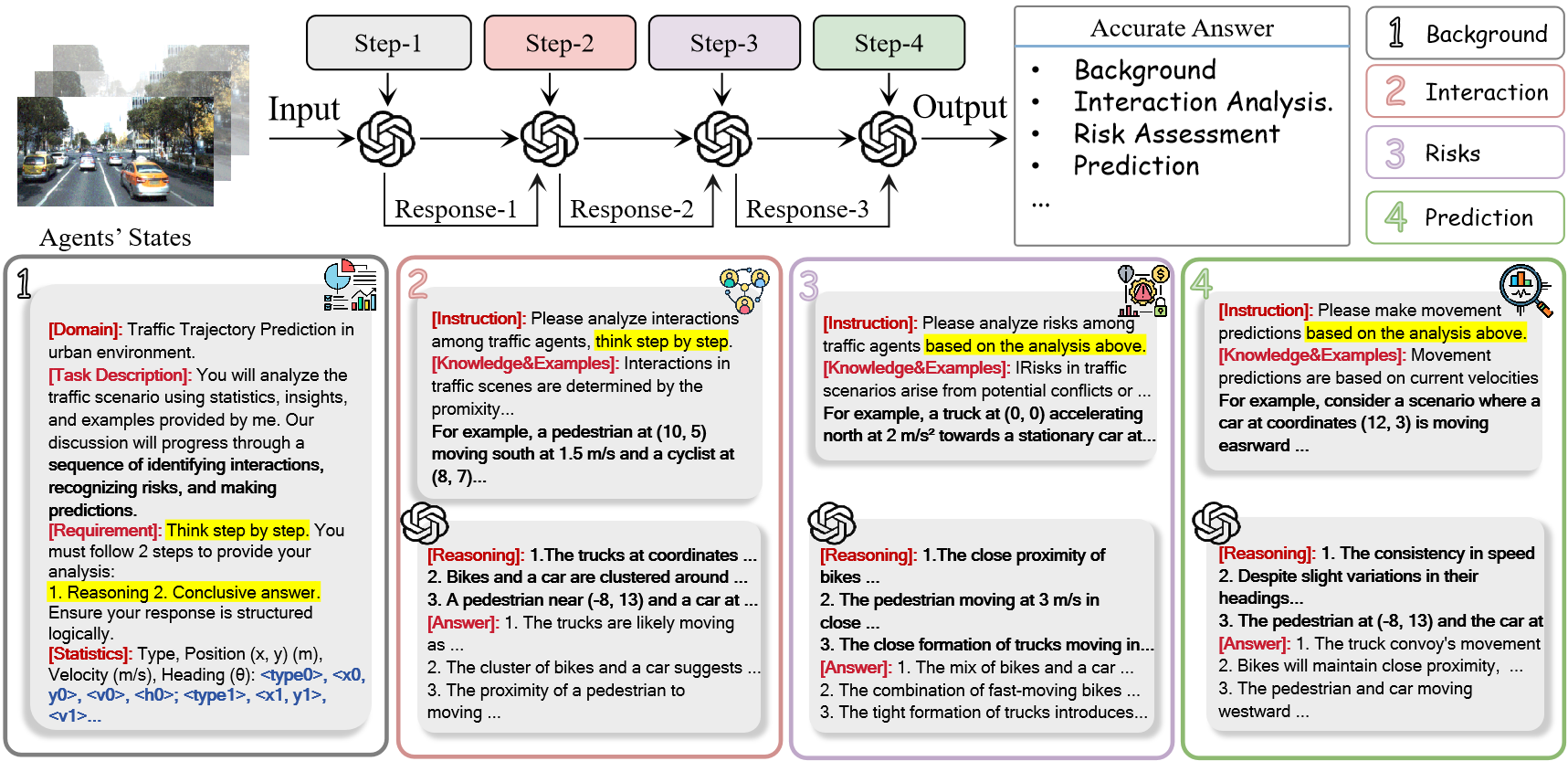}
\caption{{Illustration of the chain-of-thought prompting used in our proposed datasets to generate semantic annotations for a given traffic scene. The dialogue progression is methodically structured under human-like cognitive processes that include Background and Statistics, Interaction Analysis, Risk Assessment, and Prediction. Within each thematic category (step), we systematically infuse the LLM with specific knowledge and illustrative examples.}}
\label{fig:prompt}
\end{figure*}

\textbf{CoT Prompting in Autonomous Driving.}
CoT prompting \cite{wei2022chain} enhances the reasoning capabilities of LLMs by structuring prompts to guide step-by-step problem-solving. This technique encourages models to decompose complex tasks incrementally, mirroring human cognitive processes for clearer, more logical decision-making. By ensuring a structured reasoning flow, CoT prompting improves planning, interpretability, and overall performance in reasoning tasks. Notable studies involving CoT prompting with advanced LLMs, such as GPT-4 and Vicuna-cot \cite{zheng2023judging}, demonstrate its effectiveness in handling complex tasks.
CoT prompting has demonstrated its effectiveness in facilitating structured reasoning across various domains, including autonomous driving \cite{fu2024drive}. However, its application in AD motion forecasting remains largely unexplored. By embedding a human-like reasoning process through progressive question-answering, CoT prompting enables LLMs to perceive and respond to intricate real-world scenarios more accurately. This structured reasoning enhances the model's ability to interpret environmental cues while reducing hallucinations, ultimately improving scene comprehension and the precision of motion forecastings for AVs.}

\section{Proposed Datasets} \label{Dataset}
{ 
This study contributes to the field of motion forecasting by introducing two scene description datasets: \textit{Highway-Text} and \textit{Urban-Text}. These datasets encompass over 10 million words describing various traffic scenarios. The \textit{Highway-Text} dataset includes scene descriptions from 4,327 traffic scenarios derived from the Next Generation Simulation (NGSIM) dataset \cite{deo2018convolutional} and 2,279 scenarios from the Highway Drone Dataset (HighD) \cite{highDdataset}. Meanwhile, the \textit{Urban-Text} dataset features multi-agent scene descriptions from 3,255 samples in the Macao Connected Autonomous Driving (MoCAD) dataset \cite{liao2024bat} and 2,176 samples from ApolloScape \cite{apollo}, covering diverse environments such as campus roads, urban roads, intersections, and roundabouts. Both datasets are divided into training (70\%), validation (10\%), and testing (20\%) sets.

To enhance LLMs' comprehension of complex traffic scenes and minimize hallucinations, we develop a CoT prompting technique that uses sequential language instructions to guide LLMs step-by-step in generating context-aware semantic annotations. As illustrated in Figure \ref{fig:prompt}, CoT prompting unfolds as a progressive dialogue, with each step directing GPT-4 Turbo to focus on distinct facets of the scene. We outline the process of this CoT prompting technique as follows:

\textbf{Step-1: Background and Statistics.}
We design a unified structured prompt for both highway and urban scenes. Each data pair provides enriched information about traffic agents, including agent types, positions, velocities, headings, and environmental elements. The prompt guides LLMs in identifying key agents and generating a comprehensive overview of the current traffic situation, such as road conditions, traffic density, notable incidents, and potential behaviors of each agent.

\textbf{Step-2: Interaction Analysis.} 
This stage analyzes the interactions between traffic agents by leveraging the context from Step 1. The model assesses how agents such as vehicles, pedestrians, and cyclists influence each other, identifying key interactions likely to impact future maneuvers.

\textbf{Step-3: Risk Assessment.}  Building on the background and interaction information, this stage guides LLMs to evaluate potential accident risks. LLMs review previous findings to assess collision likelihood using factors like vehicle distribution, speed, road conditions, and pedestrian behavior. This assessment integrates risk models to calculate urgency scores based on agent type, number, and proximity, quantifying immediate risks to prioritize decision-making.

\textbf{Step-4: Prediction.}
In the final stage, the LLMs are instructed to predict the target vehicle's future maneuvers, such as acceleration, deceleration, or lane changes, and to provide justification for these predictions. In addition, the LLM generates future trajectory coordinates for the predicted maneuvers and summarises the entire reasoning process. 

\begin{figure*}[t]
\centering
\includegraphics[width=0.9\textwidth]{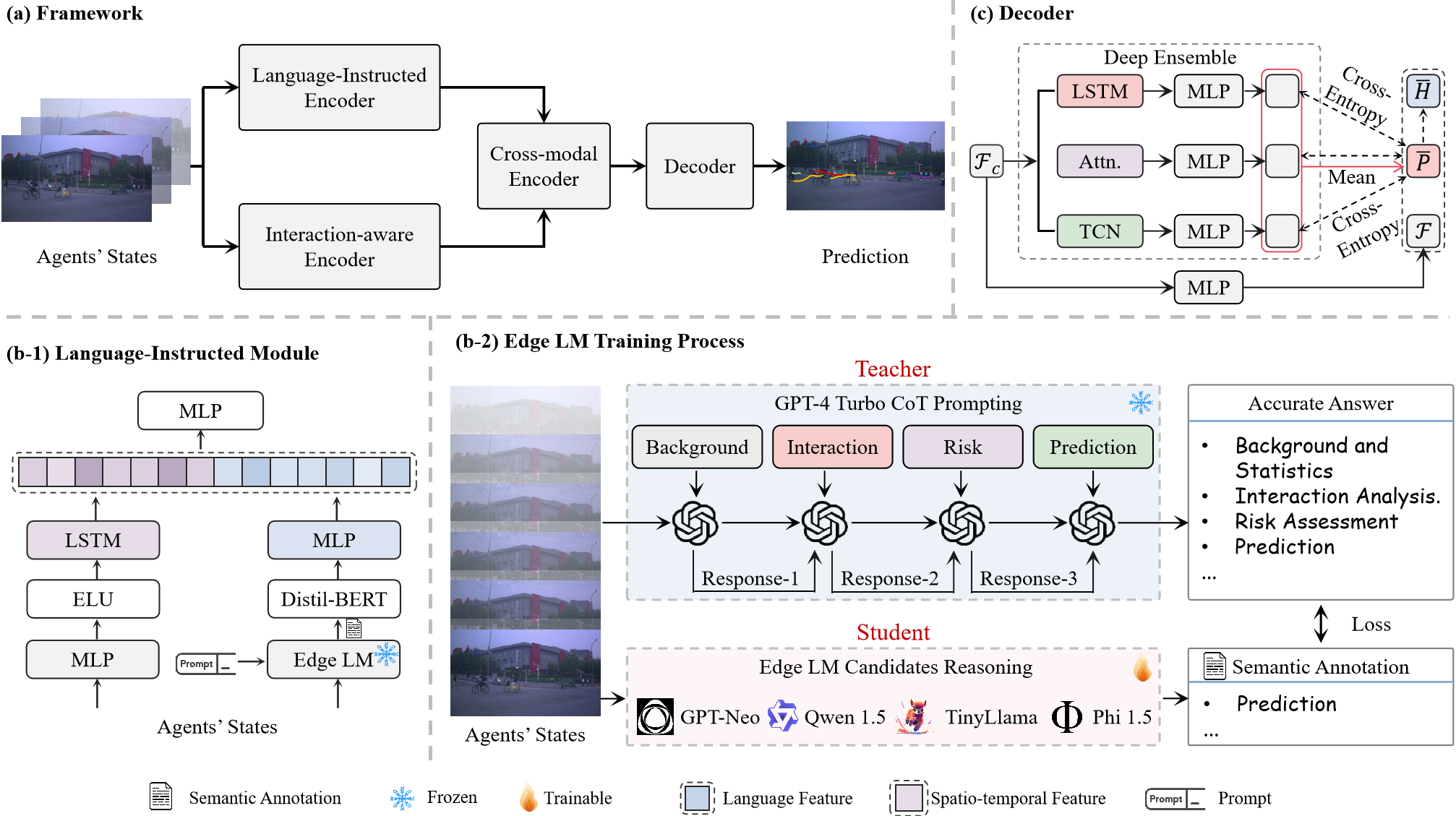}
\caption{{Overall pipeline of CoT-Drive. Panel (a) illustrates the encoder-decoder architecture of CoT-Drive, comprising four main modules: Language-Instructed Encoder, Interaction-aware Encoder, Cross-modal Encoder, and Decoder. Panels (b-1) and (b-2) illustrate the workflows of the Language-Instructed Encoder and the training process for the edge LM. This training involves multimodal fusion of semantic annotations and spatio-temporal data, where annotations are generated by a fine-tuned LM. The edge LM is trained on real-world text data labeled through CoT prompting-enhanced GPT-4 Turbo, allowing it to inherit the rich contextual learning capabilities of LLMs. Panel (c) illustrates the Decoder, which utilizes a deep ensemble method to handle aleatoric and epistemic uncertainties, combining Gaussian Mixture Models for maneuver-based predictions.}}
\label{workflow}

\end{figure*}

Through iterative refinement, insights from all four steps are synthesized into coherent semantic annotations in a standardized format. All LLM-generated annotations undergo manual validation and are cross-checked against traffic rules and legal standards to ensure compliance with the European Union's General Data Protection Regulation (GDPR) \cite{li2024steering}.
Overall, these datasets are the first to leverage GPT-4 Turbo’s linguistic capabilities with CoT prompting for detailed semantic descriptions of traffic scenarios. By introducing these datasets, we aim to advance motion forecasting models, improve generalization, and establish \textit{Highway-Text} and \textit{Urban-Text} as benchmarks for complexity and realism in AD research.}

\section{METHODOLOGY}\label{Methodology}
\subsection{Architecture Overview}
{ 
The primary objective of this study is to predict the future trajectory of the \textit{target agent} within the perceptual range of the AV. At the current time $t$, given the historical agent states $\bm{X}_{0:n}^{t-t_{h}:t}$ of the target agent (denoted with subscript 0) and its surrounding traffic agents (subscripts from 1 to $n$) over the time interval from $t-t_{h}$ to $t$, the task is to predict the future trajectory $\bm{Y}_0^{t+1:t+t_{f}}$ of the target agent over a specified prediction horizon $t_{f}$. The historical agent states $\bm{X}_{0:n}^{t-t_{h}:t}$ consists of the 2D position coordinates, heading, velocity, lane identifiers, and acceleration of the target agent and its surrounding agents.  The overall pipeline of CoT-Drive is shown in Figure \ref{workflow} (a), which is built upon the encoder-decoder paradigm, including four key components: a Language-Instructed Encoder, an Interaction-aware Encoder, a Cross-modal Encoder, and a Decoder. In a nutshell, the Language-Instructed Encoder generates semantic descriptions of the traffic scenario, including interaction analysis, risk assessment, and movement predictions, to provide a comprehensive understanding. These descriptions are then extracted as multimodal features $\mathcal{F}_{m}$, while the Interaction-aware Encoders concurrently extract localized spatial features $\mathcal{F}_{p}$. Subsequently, the Cross-modal Encoder integrates and updates the representation of these features $\mathcal{F}_{m}$ and $\mathcal{F}_{p}$ to produce the cross-modal features $\mathcal{F}_{c}$. Finally, the Decoder utilizes $\mathcal{F}_{c}$ to predict multimodal trajectories.}

\subsection{Language-Instructed Encoder}
This encoder extracts rich semantic features from complex traffic scenes, balancing accuracy and efficiency for practical use. As shown in Figure \ref{workflow}, we introduce a ``teacher-student'' knowledge distillation framework, using a pre-trained LLM, GPT-4 Turbo, as the ``teacher'' to generate semantic responses $\mathcal{A}$ for traffic scenarios based on CoT prompting. These semantic answers $\mathcal{A}$ are then used as knowledge labels to instruct a ``student'' model, a lightweight edge-optimized language model (edge LM). Under the guidance of the ``teacher'' model, the ``student'' model is fine-tuned to replicate the teacher’s capability and behavior in scene understanding and generating the semantic annotations $\mathcal{S}$. The incorporation of multimodal fusion in this encoder captures the interaction between semantic annotations $\mathcal{S}$ and historical agent states $\bm{X}_{0}^{t-t_{h}:t}$, producing multimodal features $\mathcal{F}_{m}$.

\subsubsection{Teacher Model}
To fully exploit the scene understanding capabilities of large models, we propose a novel zero-shot CoT prompting approach to guide GPT-4 Turbo in a progressive interpretation of the traffic scenarios, ultimately producing accurate answers (\(\mathcal{A}\)) for the ``student'' model. As depicted in Figure \ref{fig:prompt}, we design a series of questions (\(\mathcal{Q}\)) and prompts (\(\mathcal{T}\)) that interact with GPT-4 Turbo $p_{\mathrm{GPT}}$ in a dialogic manner, aiming to maximize the likelihood of generating accurate answers (\(\mathcal{A}\)). Mathematically,
\begin{equation}
p(\mathcal{A} \mid \mathcal{T}, \mathcal{Q})=\prod_{i=1}^{|\mathcal{A}|} p_{\mathrm{GPT}}\left(a_i \mid \mathcal{T}, \mathcal{Q}, a_{<i}\right)
\end{equation}

Here, $a_i$ and $|\mathcal{A}|$ are the $i$-th token and the length of the final answer, respectively. Then, the integration of CoT reasoning further enhances the prompts (\(\mathcal{T}\)) by embedding reasoning steps (\(\mathcal{C}\)). Formally,
\begin{equation}
p(\mathcal{A}|\mathcal{T}, \mathcal{Q}) = p(\mathcal{A}|\mathcal{T}, \mathcal{Q}, \mathcal{C}) \times p(\mathcal{C} | \mathcal{T}, \mathcal{Q})
\end{equation}
Correspondingly, $p(\mathcal{C}|\mathcal{T},\mathcal{Q})$ and $p(\mathcal{A}|\mathcal{T},\mathcal{Q}, \mathcal{C})$ are defined as:
\begin{equation}
\begin{aligned}
p(\mathcal{C}|\mathcal{T}, \mathcal{Q})& = \prod_{i=1}^{|\mathcal{C}|} p_{\mathrm{GPT}}(c_i | \mathcal{T}, \mathcal{Q}, c_{<i}) \\
p(\mathcal{A} \mid \mathcal{T}, \mathcal{Q}, \mathcal{C}) & =\prod_{j=1}^{|\mathcal{A}|} p_{\mathrm{GPT}}\left(a_j \mid \mathcal{T}, \mathcal{Q}, \mathcal{C}, a_{<j}\right)
\end{aligned}
\end{equation}
where $c_i$ is one step of total $|\mathcal{C}|$ reasoning steps. Our prompts (\(\mathcal{T}\)) are crafted to mirror human cognitive functions such as interaction-risk assessment-prediction, steering GPT-4 Turbo through a series of questions (\(\mathcal{Q}\)) that foster initial reasoning and lead to definitive answers. Moreover, each query integrates commonsense knowledge and specific examples, enabling the model to progressively refine its responses autonomously. These step-by-step CoT prompts bolster GPT’s ability to learn context and infer meanings in traffic scenarios without additional fine-tuning, thus producing precise and informative semantic answers $\mathcal{A}$ for the ``student'' model. 

\subsubsection{Student Model} 
{To reduce the computational burden during inference, we employ a lightweight edge LM as the ``student'' model to learn scene understanding capabilities from the ``teacher'' model $p_{\mathrm{GPT}}$ enhanced by CoT prompting. 
The ``student'' model takes the historical agent states $\bm{X}_{0:n}^{t-t_{h}:t}$ as inputs to produce the semantic annotations $\mathcal{S}$. Specifically, the knowledge distillation process involves the use of informative scene answers $\mathcal{A}$ to supervise the training of the student model to accurately understand the traffic scene described by $\bm{X}_{0:n}^{t-t_{h}:t}$. Formally, the learning process of the ``student" model can be defined as follows:
\begin{equation} \label{eq:kd}
\theta_\mathcal{S} \leftarrow \theta_\mathcal{S}-\eta \cdot \nabla_{\mathcal{S}} \mathcal{L}(\mathcal{S}, \mathcal{A})
\end{equation} 
where $\theta_\mathcal{S}$ is the parameter of the student model, $\eta$ denotes the learning rate and $\nabla_{\mathcal{S}} \mathcal{L}(\mathcal{S}, \mathcal{A})$ represents the gradient of the error \(\mathcal{L}(\mathcal{S}, \mathcal{A})\) between $\mathcal{S}$ and $\mathcal{A}$.
This learning process fundamentally involves the student model progressively approximating the teacher model, represented by a progressive alignment between $\mathcal{S}$ and $\mathcal{A}$. In particular, we experiment with various student models, including GPT-Neo, Qwen 1.5 \cite{bai2023qwen}, TinyLlama \cite{liusphinx}, and Phi 1.5 \cite{ali2024memory}, to investigate the impact of the parameter size on the effectiveness of knowledge distillation.} Further details are provided in Section \ref{Ablation}.

\subsubsection{Multimodal Fusion} The multimodal fusion is responsible for accepting semantic annotations $\mathcal{S}$ and the embedded historical the target agent states $\bm{X}_{0}^{t-t_{h}:t}$  and fusing them. Initially, the semantic annotations $\mathcal{S}$ undergo processing through the DistilBERT framework \cite{sanh2019distilbert}, coupled with max pooling, to extract semantic features $\mathcal{F}_s$
In parallel, the historical agent states $\bm{X}_{0}^{t-t_{h}:t}$ are fed into a Linear-ELU-LSTM network structure to generate the temporal feature $\mathcal{F}_{t}$. Finally, a Multilayer Perceptron (MLP) is used to fuse the feature of two modalities, thereby generating the multimodal features $\mathcal{F}_{m}$.

\subsection{Interaction-aware Encoder}
We employ a transformer-based structure in the encoder to capture spatial interactions between the target agent and surrounding agents. At any given time step \( t_k \in [t-t_h, t] \), the history states \( \bm{X}_{0:n}^{t_k} \) are fed into this module, which is first processed through an MLP for dimensional transformation. Then, the multi-head attention mechanisms and normalization functions are utilized to model the spatial dynamics of these representations, with shared weights across all time frames. Finally, these processed representations are passed through another MLP to generate the spatial features \( \mathcal{F}_{p} \).

\subsection{Cross-modal Encoder} 
Following the incorporation of a set of encoders, an attention mechanism is introduced prior to the decoder. This mechanism is designed to capture the cross-modal interactions of the encoded features, thereby enabling dynamic adjustment of the weights attributed to these diverse information sources. This allows the model to be tailored to meet the specific requirements of the current context. Specifically, the semantic $\mathcal{F}_{s}$, multimodal $\mathcal{F}_{m}$, and spatial $\mathcal{F}_{p}$ features are projected into query $\bar{Q}$, key $\bar{K}$, value $\bar{V}$ vectors, respectively: 
\begin{equation}
\bar{Q}=\bar{W}_{Q} \mathcal{F}_{s}, \quad
\bar{K}=\bar{W}_{K} \mathcal{F}_{m}, \quad
\bar{V}=\bar{W}_{V} \mathcal{F}_{p}
\end{equation}
where $\bar{W}_{Q}, \bar{W}_{K}, \bar{W}_{V}$ are learnable matrices. Furthermore, We make matrix product on these vectors to weight the cross-modal features as follows:
\begin{equation}
    \mathcal{F}_c=\phi_{\textit{Softmax}}(\frac{\bar{Q} \bar{K}^{\mathrm{T}}}{\sqrt{d_k}})\bar{V}
\end{equation}
Moreover, $\phi_{\textit{Softmax}}$ represents the Softmax activation function, while $d_k$ is the projection channel dimension.

\subsection{Decoder}
{ 
The decoder employs a dual-strategy to handle Aleatoric (AU) and Epistemic Uncertainty (EU) in traffic scenarios. It uses a Gaussian Mixture Model (GMM) for maneuver-based multimodal predictions, complemented by deep ensemble techniques for better adaptability to rare scenes. Maneuvers are categorized into lateral (left, right, straight) and longitudinal (accelerating, decelerating, maintaining speed) movements to model AU. Based on observed agent states \(\bm{X}_{0:n}^{t-t_{h}:t}\), the maneuver probability \(P(\bm{M}|\bm{X}_{0:n}^{t-t_{h}:t})\) is estimated, with GMM predicting future trajectories accordingly. Mathematically,
\begin{equation}
    \begin{aligned}
&P(\bm{Y}_0^{t+1:t+t_{f}}|\bm{X}_{0:n}^{t-t_{h}:t}) \\&= \sum_{\bm{M}} P_{\Lambda}(\bm{Y}_0^{t+1:t+t_{f}}|\bm{X}_{0:n}^{t-t_{h}:t}, \bm{M})  P(\bm{M}|\bm{X}_{0:n}^{t-t_{h}:t})
\end{aligned}
\end{equation}
where \(\Lambda = [\lambda_{t+1}, \lambda_{t+2}, \cdots, \lambda_{t+t_f}]\) is the Bivariate Gaussian parameters for the target agent’s future trajectory. Each \(\lambda_{t_k'} = \{ \mu_{t_k',x}, \mu_{t_k',y}, \sigma_{t_k',x}, \sigma_{t_k',y}, \rho_{t_k'}\} \) for \({t_k}' \in [t+1, t+t_f]\) includes the mean, variance, and correlation coefficient for the x- and y-positions, reflecting the uncertainty in the predicted trajectory.}
Furthermore, the deep ensemble method is utilized to model the EU. We employ \(Q\) heterogeneous models for maneuver prediction, each generating different probability distributions of maneuvers \(\bm{M}\), represented as \(P_q(\bm{M})\) for \(q \in [1, Q]\). By aggregating these model outputs, we enhance data robustness and quantify EU, as shown in Figure \ref{workflow} (c).
We calculate the ensemble's average prediction, \(\bar{P}(\bm{M})\), and measure average cross-entropy \(\bar{H}(\bm{M})\) to extract the frame-wise pyramid feature maps for these heterogeneous models. 
\begin{equation}
\bar{H}(\bm{M}) = -\sum_{q=1}^{Q} \sum_{m \in \bm{M}} \bar{P}(m) \log P_q(m)
\end{equation}
The ensemble includes diverse models like multi-layer LSTMs, Temporal Convolutional Networks (TCNs), and multi-head self-attention models, which collectively enhance the prediction of multimodal future trajectories for the target agent $\bm{Y}_{0}^{t:t+t_f}$. 
Meanwhile, \(\mathcal{F}_c\) is passed through an MLP to modify the time dimension size, changing from the past time \(t_h\) to the future time \(t_f\), resulting in \(\mathcal{F}\).
Finally, the bivariate Gaussian distribution parameters are derived using:
\begin{equation}
    \lambda_{{t_k}'} = f_{\textit{dec}}\big({\mathcal{F}_{t_k'}}\|\overline{P}(\bm{M})\|\overline{H}(\bm{M})\big)
\end{equation}
where $\|$ denotes the matrix concatenation, and \(f_{dec}\) comprises an LSTM and MLPs.

\subsection{Training and Inference}
{
\subsubsection{Training} The training process for our model is divided into two sequential stages: LMs fine-tuning and motion forecasting training. Edge LMs are fine-tuned in the first stage using our proposed Highway-Text and Urban-Text datasets. These datasets effectively encapsulate the knowledge distilled from the teacher model (GPT-4 Turbo), facilitating the learning of universal semantic scene information. This fine-tuning process follows the standard training paradigm for autoregressive language models. Considering that the scene-specific prompt and the teacher model's reference answer \(\mathcal{A}\) are merged into a complete sequence \(\mathcal{B}\) during actual training. Formally,
\begin{equation}
    \mathcal{L}_{\text{stage-1}} = 
    -\frac{1}{|\mathcal{B}|}\sum_{i=1}^{|\mathcal{B}|} \log p(b_i|b_{<i}; \theta_{\mathcal{S}})
\end{equation}
where \(b_i\) denotes the \(i\)-th token in the merged sequence \(\mathcal{B}\), \(|\mathcal{B}|\) is the length of the sequence, \(b_{<i}\) represents all tokens preceding \(b_i\), and \(\theta_{\mathcal{S}}\) represents the parameters of the student model. 
This formulation of \(\mathcal{L}_{\text{stage-1}}\) is the concrete implementation of the general error \(\mathcal{L}(\mathcal{S}, \mathcal{A})\) introduced in Equation \ref{eq:kd}, specifically tailored for autoregressive language models in the context of knowledge distillation.
This process facilitates semantic alignment at both vocabulary and semantic levels, enabling the edge LMs to internalize the teacher's reasoning patterns and contextual understanding of scene-specific information. After this stage, the edge LMs achieve superior performance in scene understanding.

Furthermore, we utilize a multitask learning strategy for the second stage loss  \( \mathcal{L}_{\text{stage-2}} \)  that includes both track prediction loss $\mathcal{L}_{\textit{traj}}$ and maneuver loss $\mathcal{L}_{\textit{mane}}$ for maneuver prediction, which can be defined as follows:
\begin{equation}
\mathcal{L}_{\text{stage-2}} = \alpha \mathcal{L}_{\textit{traj}} + (1-\alpha) \mathcal{L}_{\textit{mane}}
\label{eq:loss}
\end{equation}
where $\alpha$ is a hyperparameter. The maneuver loss $\mathcal{L}_{\textit{mane}}$ assesses the accuracy of the predicted trajectories relative to intended maneuvers:
\begin{equation}
\begin{aligned}
    \mathcal{L}_{\textit{mane}}
    &= -\sum_{m \in \bm{M}} y_m \log P(m|\bm{X}_{0}^{t-t_h:t})
\end{aligned}
\end{equation}
where $y_m$ denotes the ground truth maneuver. Moreover, the trajectory loss $\mathcal{L}_{\textit{traj}}$, according to bivariate Gaussian distribution, is defined as follows:

\vspace{-12pt}
{\begin{footnotesize}
\begin{equation}
\begin{split}
&\mathcal{L}_{\textit{traj}} =  \sum_{t_k' = t+1}^{t+t_f} \Bigg\{ \log \left( 2\pi \sigma_{t_k',x} \sigma_{t_k',y} \sqrt{1 - \rho_{t_k'}^2} \right)  + \frac{1}{2(1 - \rho_{t_k'}^2)}\\
& \left[ \frac{(\mu_{t_k',x} - x_{t_k'})^2}{\sigma_{t_k',x}^2} \right. 
\left. - \frac{(\mu_{t_k',x} - x_{t_k'})(\mu_{t_k',y} - y_{t_k'}) \rho_{t_k'}}{\sigma_{t_k',x} \sigma_{t_k',y}} \right.  \left. + \frac{(\mu_{t_k',y} - y_{t_k'})^2}{\sigma_{t_k',y}^2} \right] \Bigg\}
\end{split}
\end{equation}
\end{footnotesize}}

\vspace{-5pt}
\noindent Overall, the combined loss functions ensure that the predicted trajectories are accurate and aligned with realistic driving maneuvers, enhancing reliability in real-world conditions.

}

{ 
\subsubsection{Inference}
For the scene annotation task, historical agent states are converted into text inputs for fine-tuned edge LMs, which generate scene annotations using prompt engineering. In the motion forecasting task, the model combines semantic annotations with historical agent states to produce the multimodal future trajectory. During inference, only the knowledge-distilled lightweight LM is responsible for generating scene descriptions, thereby ensuring efficient predictions for AVs.}

\begin{figure}[htbp]
\centering
\includegraphics[width=0.5\textwidth]{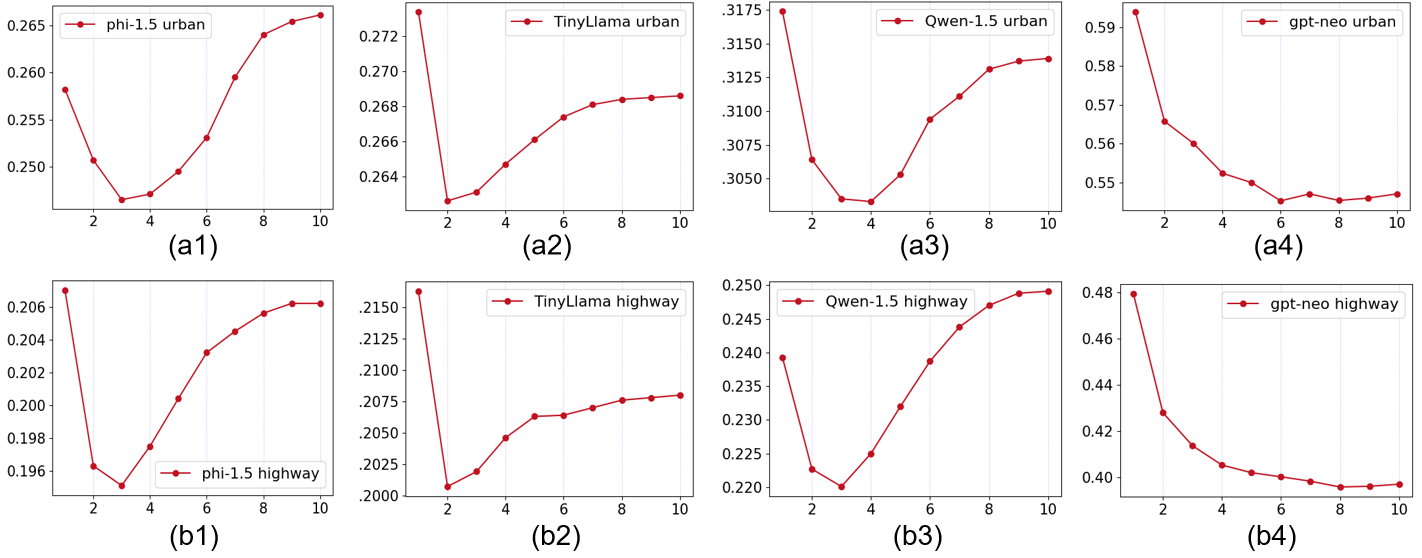}
\caption{Validation loss curves for four LMs on the developed datasets: (a1)-(a4) show the loss curves for Phi-1.5, TinyLlama, Qwen-1.5, and GPT-Neo on Urban-Text; (b1)-(b4) show their corresponding validation loss on Highway-Text dataset.}
\label{fig:train}
\end{figure}
\section{Experiments}\label{Experiment}
\subsection{Experiment Setups}
\subsubsection{Datasets}

{ 
We assess the scene understanding capabilities of the lightweight LM using the proposed Highway-Text and Urban-Text datasets. CoT-Drive's motion forecasting performance is evaluated on five real-world datasets: NGSIM, HighD, MoCAD, ApolloScape, and nuScenes.

\subsubsection{Data Segmentation}
For the NGSIM, HighD, and MoCAD datasets, trajectories are segmented into 8-second intervals: the first 3 seconds (\(t_h = 3\)) are used as input, and the following 5 seconds (\(t_f = 5\)) serve as ground truth. In ApolloScape, we follow its challenge guidelines, predicting 3-second futures (\(t_f = 3\)) based on 3-second histories (\(t_h = 3\)) for agents like vehicles, pedestrians, and cyclists. For nuScenes, we adhere to baseline methodologies, using a 2-second history (\(t_h = 2\)) to predict a 4-second future (\(t_f = 4\)).}

\subsubsection{Evaluation Metrics} 
{The effectiveness of the LM is assessed using BERT-Score \cite{hanna2021fine}, a popular metric in natural language processing to evaluate how well the generated text aligns with actual labels. This includes metrics like Precision (\(\mathbf{P}\)), Recall (\(\mathbf{R}\)), and F1 Score (\(\mathbf{F_1}\)), where values closer to 1 indicate better alignment between the LM and LLM's generative capabilities, specifically: 
\begin{equation}
  \mathbf{P} = \frac{1}{|C|} \sum_{x \in C} \max_{y \in R} \cos(\phi_\text{BERT}(x), \phi_\text{BERT}(y))
  \end{equation}
  \begin{equation}
    \mathbf{R} = \frac{1}{|R|} \sum_{y \in R} \max_{x \in C} \cos(\phi_\text{BERT}(y), \phi_\text{BERT}(x))
    \end{equation}
    \begin{equation}
      \mathbf{F_1} = 2 \times \frac{\mathbf{P} \times \mathbf{R}}{\mathbf{P} + \mathbf{R}}
\end{equation}
where $C$ represents the set of tokens in the semantic annotation $\mathcal{S}$ generated by the student LM, $R$ is the set of tokens in the accurate reference answer $\mathcal{A}$ provided by GPT-4 Turbo, and $\phi_\text{BERT}(\cdot)$ denotes the BERT embeddings for each token.

For the NGSIM, HighD, and MoCAD datasets, Root Mean Square Error (RMSE) is used as the evaluation metric. In the ApolloScape dataset, predictive accuracy is assessed using Average Displacement Error (ADE) and Final Displacement Error (FDE), with specific computations for different entities:
\begin{equation}
\begin{aligned}
\text{WSADE} &= D_v \cdot \text{ADE}_v + D_p \cdot \text{ADE}_p + D_b \cdot \text{ADE}_b, \\
\text{WSFDE} &= D_v \cdot \text{FDE}_v + D_p \cdot \text{FDE}_p + D_b \cdot \text{FDE}_b,
\end{aligned}
\end{equation}

Here, \(D_v\), \(D_p\), and \(D_b\) are the weighting factors for vehicles, pedestrians, and bicycles, respectively, set to 0.20, 0.58, and 0.22. For the nuScenes dataset, we evaluate the quality of predicted trajectories using the Minimum Average Displacement Error over \(k\) (\(\text{minADE}_k\)), Minimum Final Displacement Error over \(k\) (\(\text{minFDE}_k\)), and Miss Rate at 2 meters over \(k\) (\(\text{MR}_k\)).}

\begin{table*}[tb]
\centering
\caption{Evaluation results for our model and other SOTA baselines in the ApolloScape dataset. $\textit{ADE}_{v/p/b}$ and $\textit{FDE}_{v/p/b}$ are the ADE and FDE metrics for the vehicles, pedestrians, and bicycles, respectively. \textbf{Bold} and \underline{underlined} values represent the best and second-best performance in each category. \textcolor{blue!50}{Purple} indicates the performance of our proposed model.}
\resizebox{0.95\linewidth}{!}{
\begin{tabular}{c|c|c|c|ccc|c|ccc}
\bottomrule
Method & Publication& Backbone & WSADE & ADEv & ADEp & ADEb & WSFDE & FDEv & FDEp & FDEb \\
\hline
TrafficPredict \cite{ma2019trafficpredict} & \textcolor[RGB]{220,20,60}{AAAI'19} & LSTM & 8.5881 & 7.9467 & 7.1811 & 12.8805 & 24.2262 & 12.7757 & 11.1210 & 22.7912 \\
TPNet \cite{fang2020tpnet} & \textcolor[RGB]{220,20,60}{CVPR'20} & CNN & 1.2800 & 2.2100 & 0.7400 & 1.8500 & 2.3400 & 3.8600 & 1.4100 & 3.4000 \\
S2TNet \cite{chen2021s2tnet}  & \textcolor[RGB]{220,20,60}{ACML'21} & Transformer & 1.1679 & {1.9874} & 0.6834 & 1.7000 & 2.1798 & 3.5783 & 1.3048 & 3.2151 \\
AI-TP \cite{zhang2022ai} & \textcolor[RGB]{220,20,60}{IEEE-TIV'22} & GAT & \underline{1.1544} & 1.9878 & \underline{0.6684} & 1.6780 & {2.1297} & \underline{3.5246} & 1.2667 & 3.1370 \\
TP-EGT \cite{tp-egt} & \textcolor[RGB]{220,20,60}{IEEE-TITS'24} & Transformer & 1.1900 & 2.0500 & 0.7000 & 1.7200 & 2.1400 & 3.5300 & 1.2800 & 3.1600 \\
MSTG \cite{mstg} & \textcolor[RGB]{220,20,60}{IET-ITS'23} & LSTM & 1.1546 & \underline{1.9850} & 0.6710 & \underline{1.6745} & \underline{2.1281} & 3.5842 & \underline{1.2652} & \underline{3.0792} \\
\hline
\rowcolor{blue!8} \textbf{CoT-Drive (Ours)} & \textcolor[RGB]{220,20,60}{--} & LLM + edge LM &  \textbf{1.0958} & \textbf{1.8933} & \textbf{0.6179} & \textbf{1.6305} & \textbf{2.0260} & \textbf{3.3541} & \textbf{1.1893} & \textbf{3.0244} \\
\toprule
\end{tabular}
}
\label{tab:apollo}
\end{table*}

\subsection{Implementation Details} 
CoT-Drive is trained on four NVIDIA A100 40GB GPUs. Key implementation and parameter settings are as follows:
{ 
\subsubsection{Details of Training of LMs}
We fine-tune four LMs using $\text{bf}16$ precision and supervised fine-tuning (SFT). A learning rate of $2e^{-5}$ is selected after preliminary experiments for stability and generalization. The batch size is $8$, balancing efficiency and effective gradient updates. Training runs for $10$ epochs with a weight decay of $0.01$ to prevent overfitting. Validation loss curves (Figure \ref{fig:train}) show rapid convergence by the 10th epoch, demonstrating efficient adaptation to data distribution, reducing training costs, and facilitating faster development for real-world applications.

\subsubsection{Details of Prediction Framework}
The motion forecasting framework trains for 16 epochs with a batch size of $64$, selected for GPU memory efficiency and improved generalization. We use the Adam optimizer with Cosine Annealing Warm Restarts, adjusting the learning rate from $10^{-3}$ to $10^{-5}$ for rapid convergence and fine-tuning. The Interaction-aware Encoder uses a hidden size of $64$, with $8$ attention heads and $3$ layers, selected to balance computational efficiency and optimal validation results. The Decoder employs an ensemble of Multi-scale LSTMs, TCNs, and multihead self-attention (4 heads) to enhance robustness and accuracy. The parameter $\alpha$ in eq. \ref{eq:loss} is set to $0.5$ to balance the loss components.}

\begin{table*}[htbp]
  \centering
  \caption{Evaluation results for our proposed model and the other SOTA baselines in the NGSIM, HighD, and MoCAD datasets. Note: RMSE (m) is the evaluation metric. \textbf{Bold} and \underline{underlined} values represent the best and second-best performance.}
   \setlength{\tabcolsep}{5mm}
   \resizebox{0.9\linewidth}{!}{
    \begin{tabular}{c|cccccccc}
    \bottomrule
    \multicolumn{1}{c}{\multirow{2}[4]{*}{Datasets}}  & \multirow{2}[4]{*}{Models} & \multirow{2}[4]{*}{Backbone}& \multicolumn{6}{c}{Prediction Horizon (s)} \\
\cmidrule{4-9}    \multicolumn{1}{c}{} &  &      & 1     & 2     & 3     & 4     & 5  &AVG\\
     \hline
    \multirow{10}[3]{*}{NGSIM} 
          & CS-LSTM \cite{deo2018convolutional} & LSTM & 0.61  & 1.27  & 2.09  & 3.10  & 4.37  & 2.29  \\
            & NLS-LSTM \cite{messaoud2019non} & LSTM & 0.56  & 1.22  & 2.02  & 3.03  & 4.30  & 2.23  \\
          & MHA-LSTM \cite{messaoud2020attention} & LSTM & 0.41  & 1.01  & 1.74  & 2.67  & 3.83  & 1.93  \\  
          & WSiP \cite{Wang_Wang_Yan_Wang_2023} & LSTM & 0.56  & 1.23  & 2.05  & 3.08  & 4.34  & 2.25  \\  
          & CF-LSTM \cite{xie2021congestion} & LSTM & 0.55  & 1.10  & 1.78  & 2.73  & 3.82  & 2.00  \\  
          & TS-GAN \cite{wang2020multi} & GAN  & 0.60  & 1.24  & 1.95  & 2.78  & 3.72  & 2.06  \\  
          & STDAN \cite{chen2022intention}& Attention & 0.42  & 1.01  & 1.69  & 2.56  & 3.67  & 1.87  \\  
          & BAT \cite{liao2024bat} & LSTM &\textbf{0.23} & \textbf{0.81} & \underline{1.54} & \underline{2.52} & 3.62 & \underline{1.74} \\
          & HTPF \cite{hybrid} & LSTM & 0.49  & 1.09  & 1.78  & 2.62  & 3.65  & 1.92 \\ 
          & FHIF \cite{zuo2023trajectory} & LSTM & \underline{0.40}  & 0.98  & 1.66  & \underline{2.52}  & 3.63  & 1.84  \\  
          & \textbf{CoT-Drive (Ours)} & LLM + edge LM &  \underline{0.40}  & \underline{0.92}  & \textbf{1.43}  & \textbf{2.04}  & \textbf{2.87}  & \textbf{1.53}  \\ 
    \hline
    \multirow{7}[2]{*}{HighD} 
          & CS-LSTM \cite{deo2018convolutional}& LSTM & 0.22  & 0.61  & 1.24  & 2.10  & 3.27  & 1.49  \\ 
          & NLS-LSTM \cite{messaoud2019non}& LSTM & 0.20  & 0.57  & 1.14  & 1.90  & 2.91  & 1.34  \\ 
          & MHA-LSTM \cite{messaoud2020attention}& LSTM & 0.19  & 0.55  & 1.10  & 1.84  & 2.78  & 1.29  \\ 
          & HTPF \cite{hybrid} & LSTM & 0.16  & 0.47  & 0.94  & 1.58  & 2.36  & 1.10 \\ 
          & STDAN \cite{chen2022intention}& Attention & 0.19  & 0.27  & 0.48  & 0.91  & 1.66  & 0.70  \\ 
 
           &GaVa \cite{liao2024human}& GAT & 0.17  & 0.24  & 0.42  & 0.86  & \underline{1.31} &0.60  \\ 
          & \textbf{CoT-Drive (Ours)} & LLM + edge LM & \underline{0.08}  & \underline{0.13}  & \textbf{0.20}  & \textbf{0.37}  & \textbf{0.72}  & \textbf{0.30}  \\  
    \hline
    \multirow{7}[2]{*}{MoCAD} 
          & CS-LSTM \cite{deo2018convolutional}& LSTM & 1.45  & 1.98  & 2.94  & 3.56  & 4.49 & 2.88  \\ 
          & MHA-LSTM \cite{messaoud2020attention}& LSTM & 1.25  & 1.48  & 2.57  & 3.22  & 4.20  & 2.54  \\ 
          & NLS-LSTM \cite{messaoud2019non}& LSTM & 0.96  & 1.27  & 2.08  & 2.86  & 3.93  & 2.22  \\ 
          & CF-LSTM \cite{xie2021congestion}& LSTM & 0.72  & 0.91  & 1.73  & 2.59  & 3.44  & 1.88  \\ 
          & WSiP \cite{Wang_Wang_Yan_Wang_2023}& LSTM & 0.70  & 0.87  & 1.70  & 2.56  & 3.47  & 1.86  \\ 
          & STDAN \cite{chen2022intention}& Attention& {0.62} & {0.85} & {1.62}  & {2.51} & {3.32}  & {1.78}  \\
          & HLTP \cite{10468619} & Attention & \underline{0.55} & \underline{0.76} & \underline{1.44} & \underline{2.39} & \underline{3.21} &\underline{1.67} \\
       &  \textbf{CoT-Drive (Ours)} & LLM + edge LM & \textbf{0.43}  & \textbf{0.78}  & \textbf{1.37}  & \textbf{2.14}  & \textbf{2.79}& \textbf{1.50}  \\  
    \toprule
    \end{tabular}
    }
  \label{tab:benchmark}
\end{table*}

\subsection{Evaluations on Five Real-world Datasets}
\begin{table}[tb]
\centering
\caption{Performance comparison of models on the nuScenes dataset. Models use either HD maps (Map) and trajectory data (Traj.), or solely trajectory data ('-').}

 \setlength{\tabcolsep}{1mm}
  \resizebox{\linewidth}{!}{
\begin{tabular}{cccccc}
\bottomrule
Method & Input & Backbone & $\text{minADE}_5$ (m) & $\text{minFDE}_5$ (m) & $\text{MR}_5$ \\
  \hline
Trajectron++ \cite{salzmann2021trajectron} & Traj. + Map & Attention & 1.88 & - & 0.70 \\
MHA-JAM \cite{messaoud2021trajectory} & Traj. + Map & LSTM & 1.81 & {3.72} & - \\
EMSIN \cite{emsin} & Traj. + Map & CNN & 1.77 & \underline{3.56} & \underline{0.54} \\
AgentFormer \cite{yuan2021agentformer} & Traj. + Map & Transformer & 1.86 & 3.89 & - \\
Lapred \cite{kim2021lapred} & Traj. + Map & LSTM & \underline{1.53} & \textbf{3.37} & - \\
GOHOME \cite{gohome} & Traj. + Map & GCN & \textbf{1.42} & - & \textbf{0.57} \\
  \hline
DLow-AF \cite{yuan2020dlow} & Traj. & GRU & 2.11 & 4.70 & - \\
LDS-AF \cite{ma2021likelihood} & Traj. & LSTM  & 2.06 & 4.62 & - \\
GATraj \cite{cheng2023gatraj} & Traj. & GCN & 1.87 & 4.08 & - \\
MLST \cite{xiang2023map} & Traj. & Transformer & \underline{1.70} & \underline{3.67} & \underline{0.65} \\
AFormer-FLN \cite{10657431} & Traj. & Transformer  & 1.83 & 3.87 & - \\
\rowcolor{blue!8}\textbf{CoT-Drive (Ours)} & Traj. & LLM + edge LM & \textbf{1.56} & \textbf{3.49} & \textbf{0.52} \\
\toprule
\end{tabular}
}
\label{tab:nuscenes}
\end{table}
{ 

This study evaluates CoT-Drive against state-of-the-art (SOTA) baselines using five real-world driving datasets, with detailed results in Table \ref{tab:apollo} and Table \ref{tab:benchmark}.

\subsubsection{ApolloScape}
The ApolloScape dataset presents the challenge of predicting the trajectories of multiple agents, including vehicles, pedestrians, and cyclists, in an urban setting. CoT-Drive is built for comprehensive scene-level predictions rather than single-agent focus, enabling multi-agent motion forecastings for an entire traffic scene in a single run. As shown in Table \ref{tab:apollo}, CoT-Drive outperforms all top baselines, specifically surpassing AI-TP in WSADE by 5.1\% and MSTG in WSFDE by 4.9\%.  These findings highlight its superior ability to deliver accurate multi-agent predictions in complex urban environments, underscoring the LM’s proficiency in comprehending intricate traffic scenes.

\subsubsection{NGSIM}
The NGSIM dataset, featuring high-speed, dense highway environments, emphasizes the need for accurate prediction of vehicle intentions, such as lane changes and merging. Notably, CoT-Drive achieves a 12.07\% improvement in average prediction accuracy and a 15.59\% gain in long-term (5s) accuracy over benchmark models ranging from 2018 to 2024, highlighting its strong performance in understanding highway dynamics, in terms of long-term prediction (3s-5s).

\subsubsection{HighD}
Similar to NGSIM, the HighD dataset also involves highway traffic, but with higher trajectory data accuracy and larger sample sizes, leading to improved prediction performance. CoT-Drive demonstrates a 28.7\% increase in long-term accuracy and a 23.08\% average improvement, reinforcing CoT-Drive's leading performance in highway scenarios.

\subsubsection{MoCAD}
The MoCAD dataset contains a variety of urban scenarios, including campus streets and intersections with complex right-hand drive urban streets, which place high demands on model generalization performance. CoT-Drive consistently outperforms other models across all prediction horizons, with an average improvement of 11.33\%, underscoring its robustness in handling diverse urban conditions.

\subsubsection{nuScenes}
The nuScenes dataset includes challenging scenarios from Boston and Singapore and features high-definition (HD) maps with detailed lane and road geometry. CoT-Drive, as a map-free framework, achieves substantial improvements over non-map-based models while eliminating the need for costly HD maps. Specifically, it improves \(\text{minADE}_5\), \(\text{minFDE}_5\), and \(\text{MR}_5\) by at least 8.34\%, 5.16\%, and 2.50\%, respectively. This highlights CoT-Drive’s ability to learn contextual features of complex scenes without relying on HD maps, benefiting from CoT prompting to enhance scene understanding and multi-agent interaction. 
{\subsection{Comparison of Model Performance and Efficiency}
{

To evaluate the efficiency of CoT-Drive, we evaluate its inference speed across different configurations on the NGSIM and nuScenes datasets. Table \ref{timeforngsim} reports inference times for CoT-Drive variants (Vicuna-13B, Llama2-7B, and Mistral-7B), where each variant uses a different edge model. While CoT-Drive (Vicuna-13B) achieves the highest accuracy, the improvement is limited to 3.92\%, accompanied by a 12.7-fold increase in inference time, rendering it impractical for real-time systems. These results underscore CoT-Drive’s ability to strike an effective balance between accuracy and computational efficiency, a critical factor for real-world AVs. Furthermore, Table \ref{timefornuscenes} compares CoT-Drive with the baselines on the nuScenes dataset, evaluating predictions for 12 agents. Under identical conditions, the Llama2-7B and Vicuna-13B variants are 12 times and 20 times slower than the original CoT-Drive, while providing only marginal accuracy gains. This demonstrates that it maintains competitive inference times and significantly improves prediction accuracy, making it suitable for AV systems that require both high performance and inference efficiency.
Overall, these findings validate our model’s ability to provide efficient and accurate motion forecastings in complex environments, including highways, urban areas, and intersections, thereby addressing our first research question (\textbf{A1}).}

}

{ 
\subsection{Ablation Studies}\label{Ablation}

\subsubsection{Effect of Knowledge Distillation Strategy (A2)}  

This study employs a teacher-student knowledge distillation framework, where GPT-4 Turbo acts as the ``teacher'' to train lightweight LMs (the ``students''), enabling them to effectively understand and reason about complex driving scenarios. We validated this approach by comparing lightweight LMs such as GPT-Neo, Qwen 1.5, TinyLlama, and Phi 1.5, evaluated using our Highway-Text and Urban-Text datasets under the guidance of GPT-4 Turbo. As shown in Tables \ref{tab:text} and \ref{tab:predictive_response_time}, these models demonstrated robust scene understanding in both urban and highway contexts, reflected in high Precision, Recall, and F1 Scores, as well as competitive inference speeds, highlighting the value of our knowledge distillation strategy. Further, Table \ref{tab:module} presents the results of incorporating these distilled LMs into the Language-Instructed Encoder, demonstrating performance gains over top baselines, and indicating strong generalization across diverse traffic environments. These findings confirm that our knowledge distillation effectively transfers advanced scene understanding from GPT-4 Turbo to smaller, resource-efficient LMs, achieving accurate predictions and inference speeds fit for edge deployment. These findings directly address our second research question \textbf{(Q2)}, confirming that the knowledge distillation effectively retains the advanced scene understanding of LLMs within smaller, resource-efficient LMs.

\begin{figure}[htbp]
\centering
\includegraphics[width=0.5\textwidth]{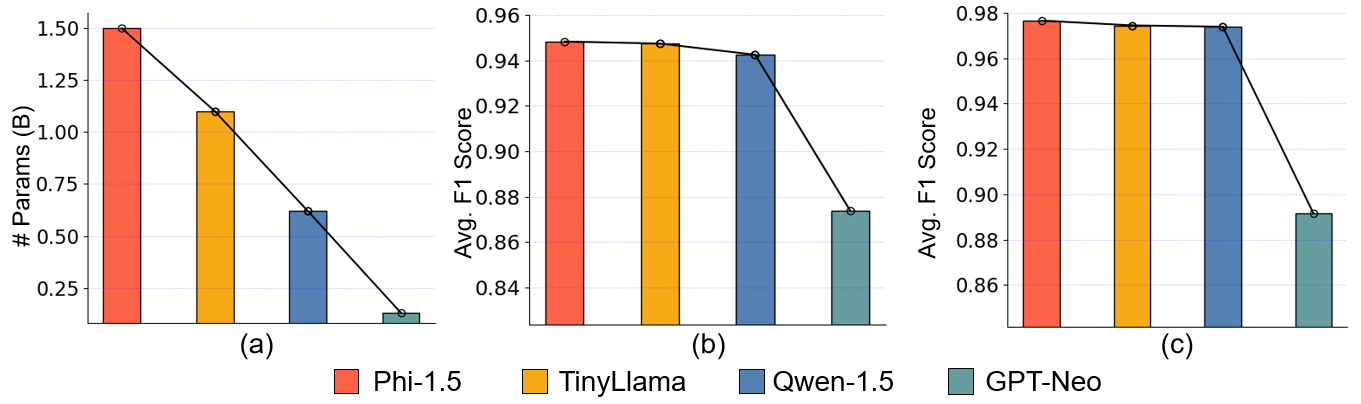}
\caption{Comparison of four different LMs in Parameter Count (a) and Performance on Urban-Text (b) and Highway-Text (c). Note: F1 Score is the evaluation metric.}
\label{fig:evaluationaa}
\end{figure}

\begin{table}[tbp]
\centering
    \caption{{Inference time comparison on NGSIM dataset for 10 batches of size 128, using Nvidia A40 GPUs.}}
 \setlength{\tabcolsep}{3mm}
\resizebox{0.85\linewidth}{!}{
\begin{tabular}{c|cc}
\bottomrule
Language Model & AVG (m) & Inference Time (s) \\
\hline

CoT-Drive & {1.53}& 0.29 \\
CoT-Drive (Vicuna-13B) & \textbf{1.47} & 2.58 \\
CoT-Drive (Llama2-7B) & 1.51& 1.92 \\
CoT-Drive (Mistral-7B) & \underline{1.50}& 1.66 \\
\toprule
\end{tabular}
}
\label{timeforngsim}
\end{table}

\begin{table}[tbp]
\centering
    \caption{{Inference time comparison on the nuScenes dataset, evaluated for 12 agents using Nvidia A40 GPUs.}}

\resizebox{1\linewidth}{!}{
\begin{tabular}{c|cccc}
\bottomrule
Language Model & minADE$_{5}$ (m)& minFDE$_{5}$ (m) & MR$_{5}$ & Inference Time (ms) \\
\hline

CoT-Drive & \underline{1.56} & \textbf{3.49} & \textbf{0.52} & 76\\
CoT-Drive (Vicuna-13B) & \textbf{1.54} & 3.60 & \underline{0.53} & 1547 \\
CoT-Drive (Llama2-7B) & 1.59 & 3.58 & 0.56 & 927\\
CoT-Drive (Mistral-7B) & 1.57 & 3.54 & 0.55 & 836 \\
\toprule
\end{tabular}
}
\label{timefornuscenes}
\end{table}

\subsubsection{Comparative Analysis of LMs}
{

Figure \ref{fig:evaluationaa} visually demonstrates that the relationship between the parameter size of LMs and their performance is not straightforwardly linear. 
To balance performance and computational burden, we conduct comparative experiments on GPT-Neo, Qwen 1.5, TinyLlama, and Phi 1.5, evaluating both their inference speeds and scene understanding capabilities. The quantitative results of the scene understanding capabilities for the LMs are detailed in Table \ref{tab:text}. We observe that, with the exception of GPT-Neo, the other three models demonstrate superior performance.
Specifically, increasing parameters from 0.13B to 0.62B yields substantial performance improvements (F1 scores rise from 0.87 and 0.89 to 0.94 and 0.97, respectively). However, further expansion from 0.62B to 1.5B produces only marginal benefits (F1 score increases of merely 0.27\% and 0.58\%). This indicates that despite a 2.4-fold increase in parameters, the performance improvement remains minimal.
To assess the impact on inference speed, we test each LM on 8 highway and 8 urban scenes, recording the time required to generate scene descriptions for all 16 scenarios. Table \ref{tab:predictive_response_time} provides a detailed comparison of inference speeds, and Figure \ref{fig:output} visualizes the semantic annotations produced by the LMs for an urban traffic scene. 
Inference time exhibits an approximately linear relationship with parameter count, ranging from 0.06 seconds for GPT-Neo to 0.37 seconds for Phi-1.5. GPT-Neo, although fastest (0.06 seconds), struggles to produce meaningful semantic annotations, whereas the other three models accurately capture complex urban dynamics, such as identifying turning and deceleration patterns of the target vehicle. Among these three, Qwen-1.5 requires the least inference time.
Based on these evaluations, \textbf{Qwen-1.5} is selected as the scene understanding component for the CoT-Drive model, offering an optimal balance between performance and computational efficiency.}

\begin{table}[htbp]
  \centering
    \caption{Performance evaluation for LMs on proposed Highway-Text and Urban-Text datasets. The evaluation metrics include Precision, Recall, and F1 Score (average across all samples). \textcolor{blue!50}{Purple} indicates the LM selected in CoT-Drive.}
  \label{tab:text}
  \resizebox{0.95\linewidth}{!}{
  \setlength{\tabcolsep}{1mm}
    \begin{tabular}{ccccccc}
    \toprule
    \multirow{3}{*}{Model}  & \multicolumn{3}{c}{Highway-Text} & \multicolumn{3}{c}{Urban-Text} \\
    \cmidrule(lr){2-4} \cmidrule(lr){5-7}
                 & Precision     & Recall   & F1 Score  & Precision     & Recall   & F1 Score \\
    \midrule

    Phi-1.5\cite{ali2024memory} &  \textbf{0.9767} & \textbf{0.9766} & \textbf{0.9766} & \textbf{0.9433} & \textbf{0.9543} & \textbf{0.9483} \\
    TinyLlama \cite{liusphinx} &   \underline{0.9739} & \underline{0.9751} & \underline{0.9745} & 0.9408 & \underline{0.9541} & \underline{0.9474} \\
    \rowcolor{blue!8}
    Qwen-1.5 \cite{bai2023qwen} &   0.9736 & 0.9745 & 0.9739 & \underline{0.9427} & 0.9426 & 0.9425 \\
    GPT-Neo \cite{nazir2023comprehensive} &   0.8899 & 0.8931 & 0.8915 & 0.8684 & 0.8792 & 0.8738 \\
    \bottomrule
    \end{tabular}
  }
\end{table}
\begin{table}[htbp]
\centering
    \caption{Inference time comparison for prediction with the batch size of $16$.  $\#$Param. is the number of parameters.}
 \setlength{\tabcolsep}{3mm}
\resizebox{0.85\linewidth}{!}{
\begin{tabular}{c|c|c}
\bottomrule
Language Model & \#Param. (B) & Inference Time (s) \\
\hline
GPT-Neo & 0.13 & 0.06 \\
\rowcolor{blue!8}
Qwen-1.5 & 0.62 & 0.17 \\
TinyLlama & 1.10& 0.29 \\
Phi-1.5 & 1.50& 0.37 \\
\toprule
\end{tabular}
}
\label{tab:predictive_response_time}
\end{table}

\subsubsection{Importance of Each Component}

\begin{table}[htbp]
\centering
\caption{Evaluation of different ablated models, where LI, IA, CA, and UD denote the Language-Instructed Encoder, Interaction-aware Encoder, Cross-modal Attention, and Decoder, respectively. AVG denotes the average RMSE metric.}
\resizebox{0.9\linewidth}{!}{
  \setlength{\tabcolsep}{8pt}
\begin{tabular}{cccccl}
\bottomrule
\multirow{3}{*}{Ablated Models} & \multicolumn{4}{c}{Components} & \multirow{3}{*}{AVG}\\
\cmidrule(lr){2-5}
 & LI & IA & CA & UD\\
\hline
A & \textcolor[RGB]{220,20,60}{\ding{55}} & \textcolor[RGB]{34,139,34}{\ding{51}} & \textcolor[RGB]{34,139,34}{\ding{51}} & \textcolor[RGB]{34,139,34}{\ding{51}} & 1.97$_{\downarrow{22.34\%}}$ \\
B & \textcolor[RGB]{34,139,34}{\ding{51}} & \textcolor[RGB]{220,20,60}{\ding{55}} & \textcolor[RGB]{34,139,34}{\ding{51}} & \textcolor[RGB]{34,139,34}{\ding{51}} &1.82$_{\downarrow{15.93\%}}$  \\
C & \textcolor[RGB]{34,139,34}{\ding{51}} & \textcolor[RGB]{34,139,34}{\ding{51}} & \textcolor[RGB]{220,20,60}{\ding{55}} & \textcolor[RGB]{34,139,34}{\ding{51}} & 1.69$_{\downarrow{9.47\%}}$ \\
D & \textcolor[RGB]{34,139,34}{\ding{51}} & \textcolor[RGB]{34,139,34}{\ding{51}} & \textcolor[RGB]{34,139,34}{\ding{51}} &  \textcolor[RGB]{220,20,60}{\ding{55}} & 1.88$_{\downarrow{18.62\%}}$ \\
E & \textcolor[RGB]{34,139,34}{\ding{51}} & \textcolor[RGB]{34,139,34}{\ding{51}} & \textcolor[RGB]{34,139,34}{\ding{51}} & \textcolor[RGB]{34,139,34}{\ding{51}} &1.53  \\
\toprule
\end{tabular}
}
\label{tab:module}
\end{table}

To assess the importance and effectiveness of each component in CoT-Drive, we conduct extensive ablation studies focusing on four key components of CoT-Drive, as summarized in Table \ref{tab:module}.  Model A omits the knowledge distillation step in the Language-Instructed Encoder, using LMs directly for scene descriptions instead. This results in a significant performance drop of up to 22.34\%, highlighting the crucial role of knowledge distillation in enabling LMs to inherit advanced scene understanding capabilities from LLMs—especially in complex environments involving multi-agent interactions, congested streets, and intersections. These findings directly address our first research question (\textbf{Q1}), underscoring that knowledge distillation is vital for transferring sophisticated reasoning skills to lightweight LMs while maintaining computational efficiency. 

Model B omits the Interaction-aware Encoder, leading to a 15.93\% performance drop, emphasizing the need to capture both scene semantics and spatio-temporal interactions for effective multi-agent context handling. Model C replaces the cross-attention mechanism with an MLP, resulting in a 9.47\% decrease, underscoring cross-attention's role in accurately focusing on key traffic agents and environmental cues. Model D substitutes the proposed decoder with a simple MLP, causing an 18.62\% performance drop, likely due to its inability to handle traffic uncertainties. The original decoder uses maneuver-based multi-modal prediction to manage unpredictability, highlighting the need for explicitly modeling scene uncertainty to ensure reliability.

\begin{figure}[htbp]
\centering
\includegraphics[width=0.48\textwidth]{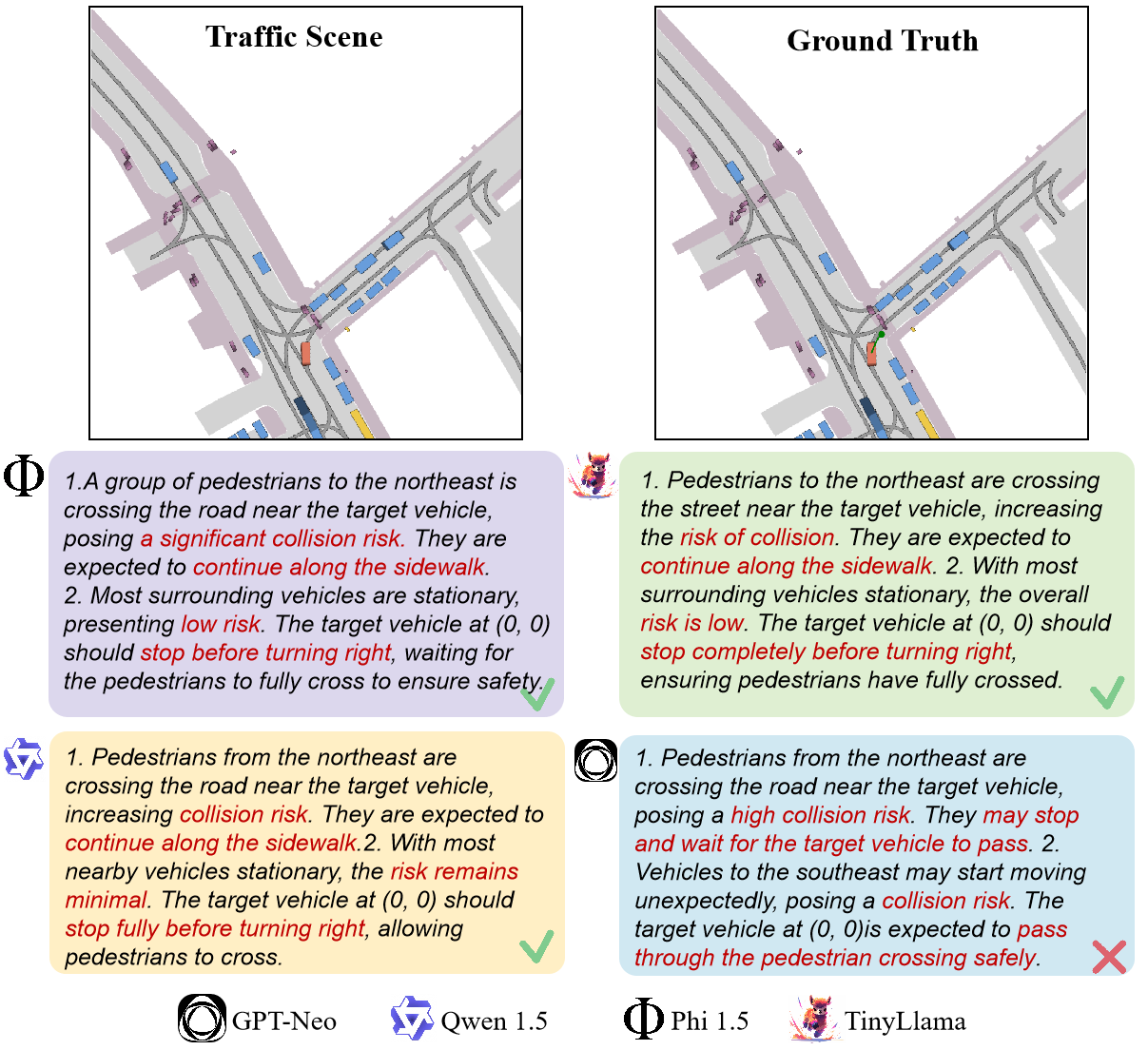}
\caption{Comparison of semantic annotation output by four different LMs for a specific traffic scene in the nuScenes dataset. In the ``Traffic Scene'' subfigure, the target agent is marked in red, while the ground truth trajectory in the ``Ground Truth'' subfigure is depicted in green.}
\label{fig:output}
\end{figure}

\subsubsection{Ablation Studies on the CoT Prompting (A3)}
We further investigate the impact of CoT prompting and the text-generation capabilities of various LMs on motion forecasting performance. Table \ref{tab:lm} presents a detailed ablation analysis of different LMs and the effects of CoT prompting. Results reveal that the three LMs, which performed similarly well in generating scene annotations on the CoT prompting-based dataset, also demonstrated comparable prediction performance with minimal differences. Even the smallest LM maintained competitive performance, highlighting its efficiency within our framework. However, when CoT prompting is removed from the dataset generation process, significant declines in prediction performance were observed across all LMs, regardless of their parameter size. This underscores the pivotal role of CoT prompting in improving model performance by guiding LLMs to provide deep and nuanced scene descriptions by eliciting a step-by-step reasoning process. The enriched scene annotations generated by the CoT prompting-based LLMs, in turn, serve as a solid foundation for embedding complex semantic understanding into lightweight LMs, ultimately enhancing the inference capabilities of CoT-Drive. These findings also directly answer our second research question (\textbf{Q3}), offering strong evidence that CoT prompting significantly enhances the contextual understanding of LLMs in complex traffic scenarios without requiring additional fine-tuning.

\begin{table}[htbp]
\centering
\caption{Ablation study results for different LMs that were finetuned on datasets annotated with and without CoT prompting, used in Language-Instructed Encoder.}
\resizebox{0.95\linewidth}{!}{
\begin{tabular}{c|c|c|cc}
\bottomrule
Model & \# params (B) & CoT & WSADE (m) &  WSFDE (m) \\
\hline
 Phi-1.5 & 1.50 & \textcolor[RGB]{220,20,60}{\ding{55}} &1.2076 & 2.2139 \\
 {Phi-1.5} & {1.50} & \textcolor[RGB]{34,139,34}{\ding{51}} &{1.0924} & {2.0191} \\
  TinyLlama & 1.10 & \textcolor[RGB]{220,20,60}{\ding{55}} & 1.1923 & 2.2086 \\
   TinyLlama & 1.10 & \textcolor[RGB]{34,139,34}{\ding{51}} & 1.0931 & 2.0203 \\
     Qwen-1.5 & 0.62 &\textcolor[RGB]{220,20,60}{\ding{55}} & 1.1874 & 2.1903 \\
\rowcolor{blue!8}
 Qwen-1.5 & 0.62 &\textcolor[RGB]{34,139,34}{\ding{51}} & 1.0958 & 2.0260 \\
  GPT-Neo & 0.13 &\textcolor[RGB]{220,20,60}{\ding{55}} & 1.2743 & 2.3201 \\
 GPT-Neo & 0.13 &\textcolor[RGB]{34,139,34}{\ding{51}} & 1.1546 & 2.1284 \\
\toprule
\end{tabular}
}
\label{tab:lm}
\end{table}

\subsubsection{Ablation Studies on the Decoder}
We explore the influence of different ensemble methods within the decoder, examining the influence of different architectures and the number of submodels on the results, as shown in Table \ref{tab:ensemble}. The results indicate that omitting all sub-models leads to a performance drop of 8.3\% in CoT-Drive's predictions. Gradually increasing the number of sub-models shows a marked improvement in prediction accuracy, highlighting the effectiveness of our proposed dual-strategy framework for handling AU and EU in traffic scenarios. Interestingly,  increasing the number of sub-models leads to diminishing returns beyond a certain point. Specifically, the six sub-model configurations combining MS-LSTM, self-attention, and TCN provide optimal performance, even outperforming the nine sub-model configurations. This suggests that excessive model complexity can lead to overfitting or redundancy, highlighting the importance of balancing model complexity and efficiency to ensure robust motion forecasting in diverse traffic environments.

}

\begin{table}[htp]
\centering
\caption{Ablation study results for different numbers of ensembled models within the Decoder. AVG: average RMSE metric. Self-Attn: self-attention mechanism.}
\resizebox{0.8\linewidth}{!}{
  \setlength{\tabcolsep}{7pt}
\begin{tabular}{ccccc}
\bottomrule
\multicolumn{3}{c}{Architectures} & \multirow{3}{*}{Sub-Models} & \multirow{3}{*}{AVG} \\
\cmidrule(lr){1-3}
MS-LSTM & Self-Attn. & TCN  \\
\midrule
\textcolor[RGB]{220,20,60}{\ding{55}} & \textcolor[RGB]{220,20,60}{\ding{55}} & \textcolor[RGB]{220,20,60}{\ding{55}} & 0 & 1.67 \\
\textcolor[RGB]{34,139,34}{\ding{51}} & \textcolor[RGB]{220,20,60}{\ding{55}} & \textcolor[RGB]{220,20,60}{\ding{55}} & 1 & 1.61 \\
\textcolor[RGB]{34,139,34}{\ding{51}} & \textcolor[RGB]{34,139,34}{\ding{51}} & \textcolor[RGB]{220,20,60}{\ding{55}} & 2 & 1.57 \\
\textcolor[RGB]{34,139,34}{\ding{51}} & \textcolor[RGB]{34,139,34}{\ding{51}} & \textcolor[RGB]{34,139,34}{\ding{51}} & 3 & 1.54 \\
\textcolor[RGB]{34,139,34}{\ding{51}} & \textcolor[RGB]{34,139,34}{\ding{51}} & \textcolor[RGB]{34,139,34}{\ding{51}} & \textbf{3$\times$2} & \textbf{1.53} 
\\
\textcolor[RGB]{34,139,34}{\ding{51}} & \textcolor[RGB]{34,139,34}{\ding{51}} & \textcolor[RGB]{34,139,34}{\ding{51}} & 3$\times$3 & 1.56\\
\toprule
\end{tabular}
}
\label{tab:ensemble}
\end{table}

\begin{figure}[htbp]
\centering
\includegraphics[width=0.48\textwidth]{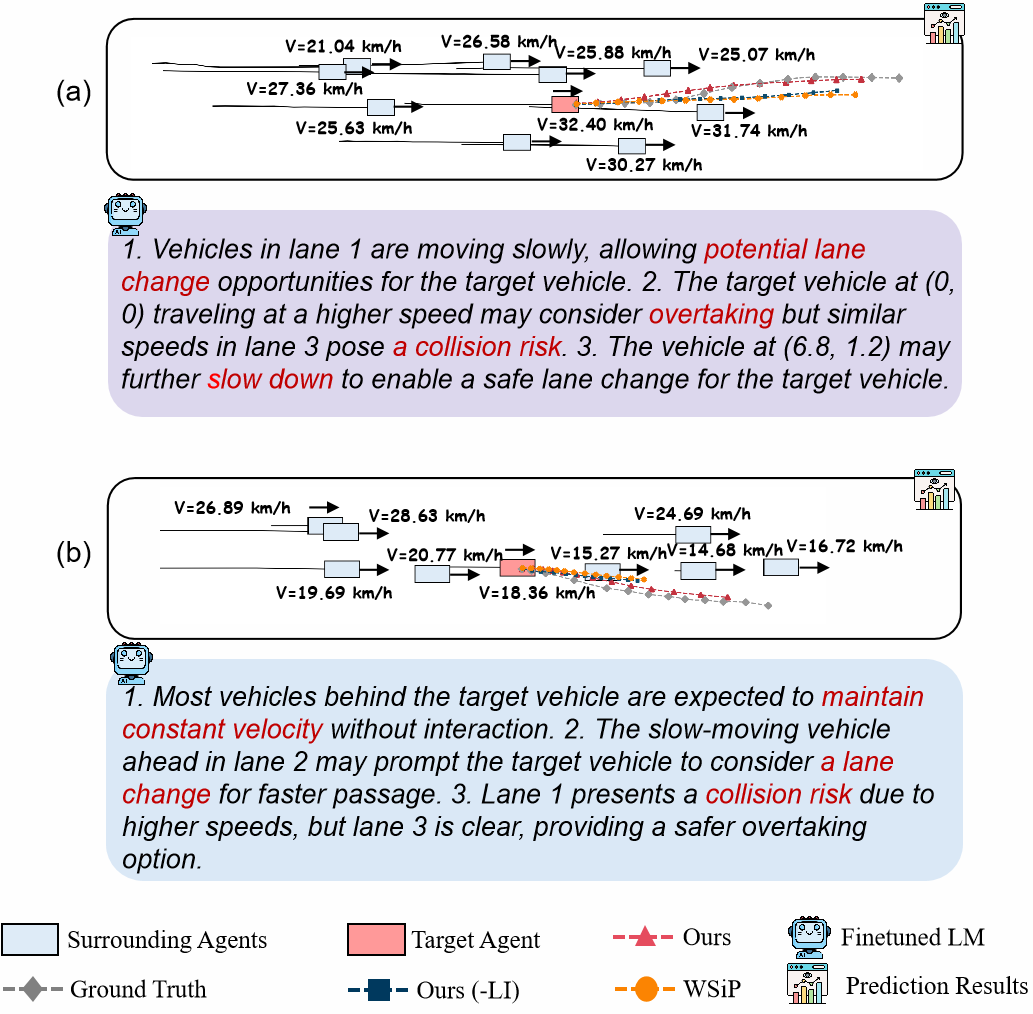}
\caption{Qualitative results compare the CoT-Drive (Ours) on the NGSIM dataset against its variant without the Language-Instructed Encoder (Ours (-LI)) and WSip.}
\label{fig:visualization1}
\end{figure}
\begin{figure*}[tb]
\centering
\includegraphics[width=0.9\textwidth]{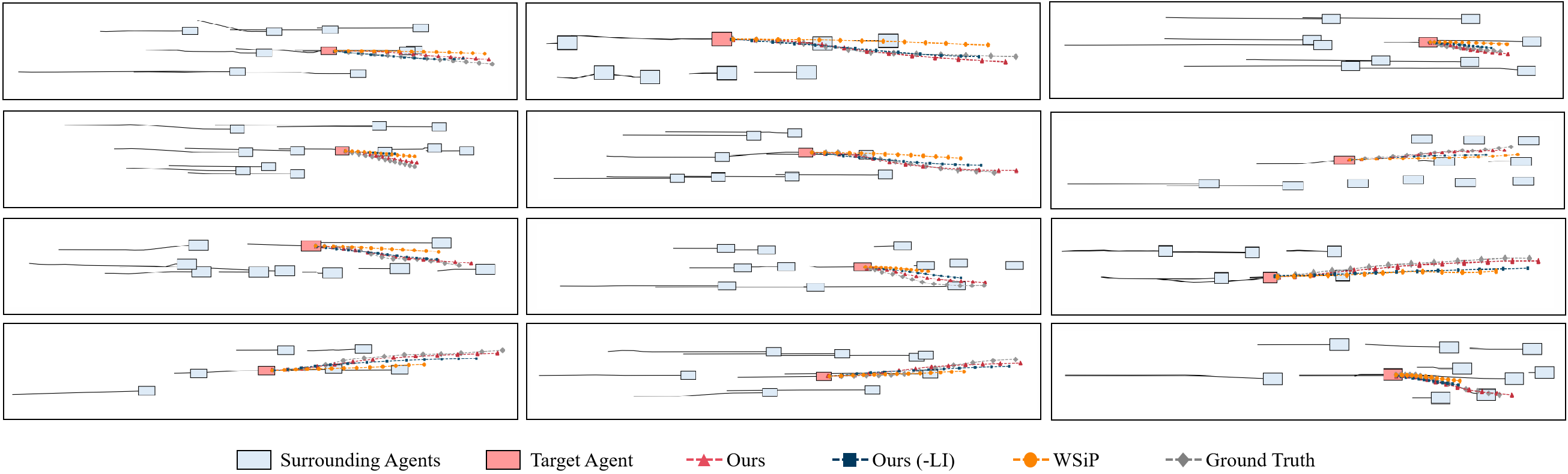}
\caption{{ Qualitative results of lane changing, merging, and acceleration scenarios on the NGSIM dataset, showcasing the effectiveness of CoT-Drive versus its variant without the Language-Instructed Encoder (Ours (-LI)) and the baseline (WSiP).}}
\label{fig:vis}
\end{figure*}

\begin{figure}[htbp]
\centering
\includegraphics[width=0.48\textwidth]{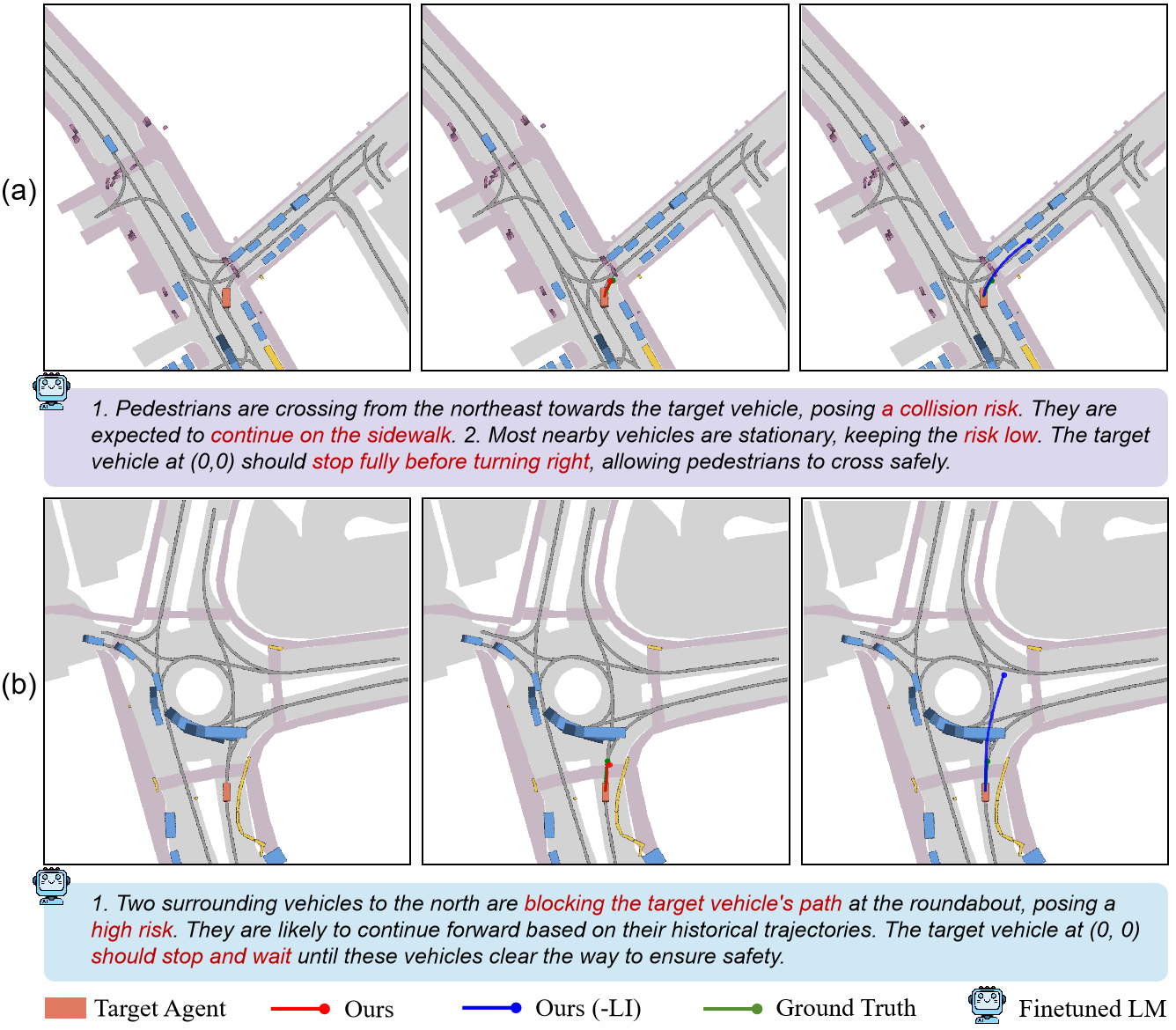}
\caption{Qualitative results compare the CoT-Drive on the nuScenes dataset (without using the HD map information) against its variant without the Language-Instructed Encoder.}
\label{fig:visualization2}
\end{figure}

\subsection{Qualitative Results}\label{Quantitative}
{ 

We present qualitative visualizations of CoT-Drive's performance in challenging highway and urban scenes, comparing it with top baselines. These visualizations highlight how CoT-Drive handles complex traffic scenes, supporting our responses to research questions (\textbf{Q1-Q3}).

\subsubsection{Qualitative Results in Highway Scenes}
Figure \ref{fig:visualization1} compares CoT-Drive, its variant without the Language-Instructed Encoder (Ours (-LI)), and the top baseline WSiP in complex highway scenarios. The fine-tuned LM correctly identifies the target vehicle’s intent to lane-change by analyzing the speed difference with the vehicle ahead, predicting an overtaking maneuver. This insight allows CoT-Drive to correctly predict the lane change, whereas the baseline WSiP and Ours (-LI) both mistakenly predict the vehicle will continue straight. These results emphasize the critical role of the Language-Instructed Encoder in capturing nuanced scene details and improving prediction accuracy. Additionally, in Figure \ref{fig:visualization1} (b), CoT prompting strengthens the model’s performance by generating detailed semantic annotations from both interaction and risk perspectives. Figure \ref{fig:vis} further illustrates CoT-Drive’s superior performance in challenging scenes, showcasing the improvements brought by the Language-Instructed Encoder and responding to the \textbf{Q1-Q2}.
}

\begin{figure*}[t]
\centering
\includegraphics[width=0.9\textwidth]{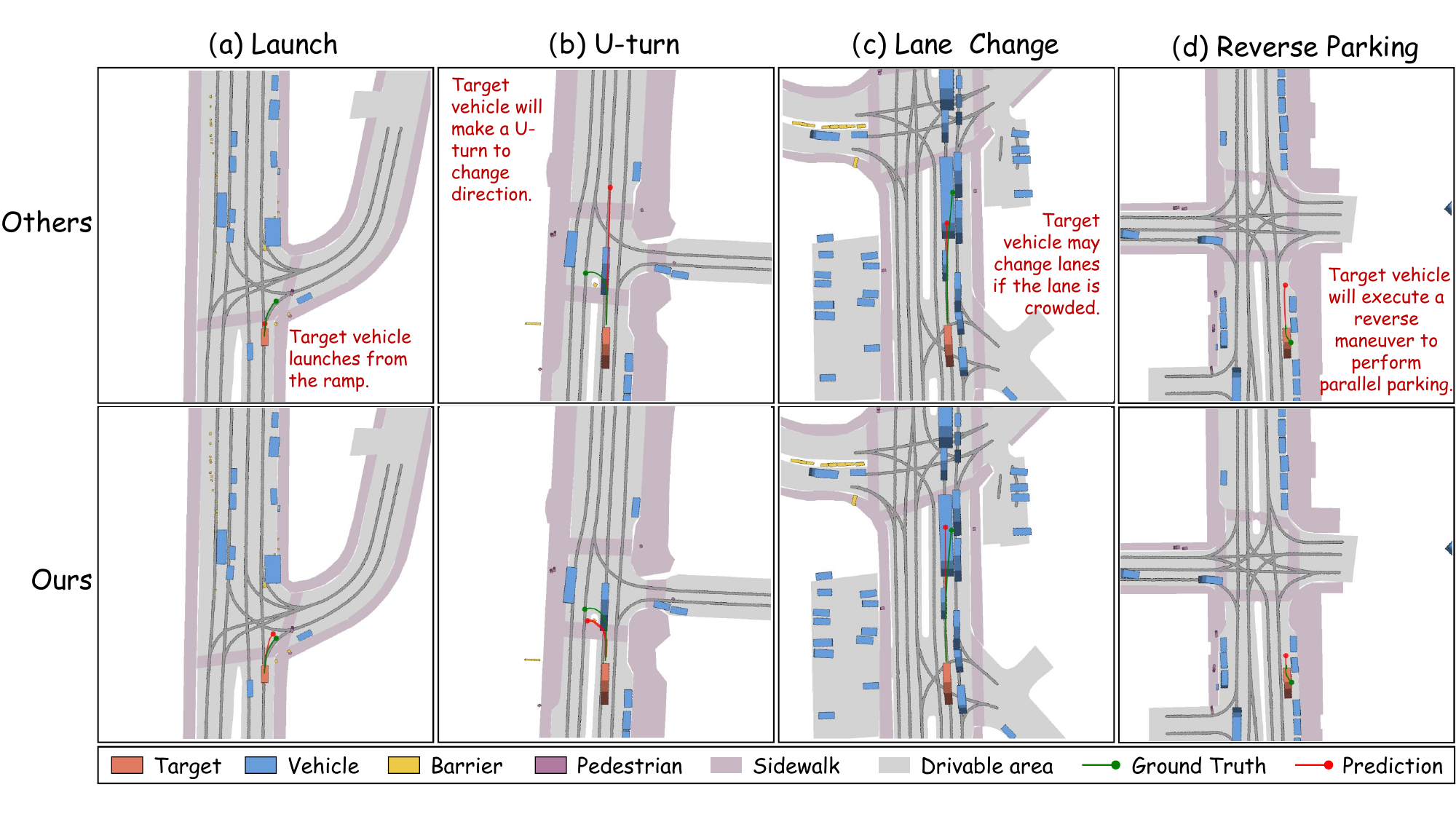}
\caption{{ Failure cases of trajectories predicted by CoT-Drive and baseline in rare and challenging scenarios from the nuScenes dataset, including launch (a), U-turn (b), lane changing (c), and reverse parking (d) scenes. For each scenario, we visualize the environment around the target vehicle, along with the {{predicted trajectory}} (Others \cite{salzmann2021trajectron} and Ours) and the {ground truth}.}}
\label{corner-case}
\end{figure*}

{ 
\subsubsection{Qualitative Results in Complex Urban Scenes}
Figure \ref{fig:visualization2} shows results from complex urban scenes in nuScenes dataset, comparing CoT-Drive with its variant without the Language-Instructed Encoder ((Ours (-LI))), supported by scene descriptions from the LM. For clarity, only one predicted trajectory is shown for each model.
Interestingly, without explicit traffic light states or high-definition (HD) maps, CoT-Drive—leveraging its fine-tuned LM—accurately infers the target vehicle's behavior to a changing signal. As illustrated in Figure \ref{fig:visualization2} (a), CoT-Drive predicts deceleration and stopping, while CoT-Drive (-LI) mistakenly predicts a turn. This highlights the Language-Instructed Encoder’s role in enhancing CoT-Drive’s human-like reasoning through CoT prompting and context-aware scene descriptions. In Figure \ref{fig:visualization2} (b), we show a scenario with a left turn by the target vehicle. CoT-Drive, using detailed scene understanding from the Language-Instructed Encoder, considers pedestrians crossing and vehicles moving straight, predicting the vehicle will decelerate to avoid collisions before turning. In contrast, CoT-Drive (-LI) incorrectly predicts an immediate left turn, overlooking key surrounding interactions. These visualizations clearly demonstrate the advantage of the step-by-step reasoning approach, resembling human cognition. The Language-Instructed Encoder enables CoT-Drive to excel in contextual understanding, accurately predicting agent intentions, and making informed decisions. These results confirm the effectiveness of CoT-Drive, demonstrating its real-world applicability for AD systems (\textbf{Q1-Q3}).}

{ 
\section{Discussions} \label{Discussion}
\subsection{Contribution Summary}
In light of the insights gained from evaluations and analysis, we summarize the primary contributions of this study:

1) This study introduces the Language-Instructed Encoder that uses \textbf{knowledge distillation} to equip a lightweight language model with the advanced scene understanding capabilities of GPT-4 Turbo. This enhancement enables real-time operation on in-vehicle edge devices, significantly improving generalization abilities. To the best of our knowledge, this is the first time a lightweight surrogate model of an LLM has been trained and used for motion forecasting tasks.

2) This study presents two innovative \textbf{scene description datasets}, \textit{Highway-Text} and \textit{Urban-Text}, derived from real-world data and enriched using CoT prompting with GPT-4 Turbo. These datasets are pioneering in utilizing LLMs for semantic understanding in motion forecasting, providing a valuable resource for further research and development in the fields of AD and LLMs.

3) Through extensive testing across the NGSIM, HighD, MoCAD, ApolloScape, and nuScenes datasets, CoT-Drive demonstrates superior performance, surpassing most SOTA baselines. This highlights its remarkable accuracy and effectiveness in various traffic scenes, including \textbf{highways}, \textbf{intersections}, and \textbf{dense urban areas}.

\subsection{Discussion of Limitations and Future Researches}
Despite the remarkable advancements demonstrated by CoT-Drive in terms of prediction accuracy and efficiency, several key challenges remain that require further exploration.

\subsubsection{Robustness in Rare and Challenging Scenarios}

We visualize several ``corner-case'' scenarios to analyze the limitations of CoT-Drive and to indicate future improvements. In Figure \ref{corner-case} (a), when a target agent starts from a complete stop, the extended stationary state introduces significant uncertainty. While CoT-Drive's predicted trajectory is closer to the ground truth than the baseline, it also struggles to infer the agent's intent, often predicting continued stationary behavior, leading to inaccuracies in multi-modal trajectories. In Figures \ref{corner-case} (b) and (c), congested lane-changing and U-turn situations require precise identification of interactions between agents to predict appropriate decisions. CoT-Drive, like other models, fails to predict these rapid behavioral shifts. Motion Forecasting models typically rely on extrapolating established trends, but in the reverse parking scenario (Figure \ref{corner-case} (d)), the target vehicle unexpectedly begins a backward maneuver, which deviates from the previously observed trajectory patterns. Here, both the CoT-Drive and the baseline lack sufficient data to predict this shift accurately, leading to errors in the prediction. However, CoT-Drive's trajectory, supported by the LMs' scene understanding, remains closer to the ground truth than other models, emphasizing the value of using linguistic intermediate representations for action reasoning with language priors. These scenarios, although rare in the real world, are essential to ensure the safe deployment of AVs, highlighting the need for future work to develop methods that use semantic information to better capture multimodal driving behaviors and accurately understand agent intentions in different scenes—not just relying on numerical data.

\subsubsection{Data Dependency of LLMs in Real-World Scenarios} 
Another limitation of CoT-Drive is its reliance on high-quality, well-annotated datasets, which are challenging and costly to obtain, particularly for rare or complex scenarios such as congested streets and adverse weather. Data collection is further complicated by privacy concerns, biases, and the need to represent a wide range of driving conditions. As a result, CoT-Drive, similar to other LLM-based models, faces challenges in adapting to unseen or rare scenarios. While CoT prompting enhances scene understanding and reduces hallucinations without the need for fine-tuning, its effectiveness is tightly coupled with the quality of the training data. Low-quality data can result in less accurate predictions. To overcome these challenges, future research could focus on generating synthetic datasets through simulations, using domain adaptation techniques to bridge the gap between simulated and real-world data. Additionally, unsupervised or semi-supervised learning approaches could reduce reliance on large annotated datasets, while continual learning mechanisms could enable CoT-Drive to dynamically adapt to evolving traffic patterns and conditions. These improvements would mitigate the current limitations and pave the way for more adaptable, reliable, and robust motion forecasting models for AD.

\subsection{Discussion of Unsupervised and Semi-supervised Approaches for Scene Description Generation}
To address the data demands of LLM-based models, unsupervised and semi-supervised learning approaches offer promising alternatives \cite{chen2022semi}. Unsupervised learning can learn from large volumes of unlabeled data, autonomously identifying patterns, thus achieving broader generalization without expensive manual labeling. Semi-supervised learning uses labeled seed data to generate pseudo-labels for unlabeled data. While these approaches reduce the need for annotated datasets, they also pose significant challenges, especially given the complexity and uncertainty of AD: 1) Data Quality: They depend on inherent data patterns, which may be insufficient for complex driving scenes, leading to unreliable predictions. In contrast, CoT-Drive uses LLM-derived semantic annotations to achieve richer contextual understanding. 2) Interpretability: They often lack explicit scene representations, reducing interpretability and hindering the validation of predictions. CoT-Drive, enhanced by CoT prompting, provides step-by-step reasoning and clear explanations, improving decision transparency. While these approaches reduce data requirements, they lack interpretability and semantic depth compared to CoT-Drive. Future research could integrate LLM-based methods with unsupervised or semi-supervised techniques to balance scalability with advanced understanding and decision-making.

\subsection{Discussion of Reinforcement Learning and Evolutionary Learning Approaches for Motion Forecasting}
This subsection compares LLM-based, RL-based, and EL-based methods across three dimensions:
\subsubsection{Data Requirements and Generalization} LLM-based methods rely on curated datasets with high-quality and labor-intensive annotations. In contrast, RL and EL methods \cite{fajardo2011automated} leverage simulated environments with dynamic obstacles, minimizing data preparation and enabling broader exploration of varied conditions. However, the ``reality gap''—the discrepancy between simulation and real-world conditions—limits the generalizability of RL/EL, due to the difficulty in designing robust reward functions for diverse situations.

\subsubsection{Training Time and Computational Complexity} LLMs involve extensive training time and computational resources due to their large model size. RL and EL approaches, though requiring smaller models and thereby reducing computational costs, often depend on extensive trial-and-error processes to converge, especially in complex reward structures. CoT-enhanced CoT-Drive provides comprehensive scene understanding without additional fine-turn, and through knowledge distillation, which makes LLM functionalities feasible for lightweight, edge-level deployment in real-world driving.

\subsubsection{Scene Understanding and Reasoning Capabilities} LLMs excel at providing rich, context-aware reasoning, offering a nuanced understanding of traffic scenes. By contrast, RL and EL models, primarily optimize policies, which often lack the semantic depth and adaptability of LLMs. RL and EL approaches typically focus on learned behaviors without interpreting broader environmental cues or explaining decision-making processes, limiting their efficacy in complex, unpredictable situations. In comparison, LLM-based methods adapt decisions based on agent intention, behaviors, and environmental context, while also providing explicit reasoning for these decisions. This makes LLM-based approaches particularly suited for scenes that require context-specific reasoning.

\subsection{Why LLMs Are a Viable Long-Term Alternative?}
LLM-based approaches offer a promising long-term solution for AD due to their advanced scene understanding and context-aware reasoning, which surpasses the capabilities of unsupervised, semi-supervised, and reinforcement learning models. Despite the need for well-curated data, LLMs excel in interpreting complex real-world environments, supported by techniques like CoT prompting that emulate human-like reasoning \cite{radford2021learning}. Unlike RL and EL models, which are computationally efficient but often lack semantic depth, LLMs integrate diverse pre-training and multi-source data, enabling nuanced understanding. Advancements in knowledge distillation and synthetic data generation mitigate extensive training requirements, facilitating edge-level deployment and reducing dependency on large datasets. With improvements in onboard computational power, deploying sophisticated LLMs in AVs becomes more feasible, enhancing both real-time performance and reliability \cite{NEURIPS2020_1457c0d6}. Overall, LLMs strategically balance computational efficiency with predictive capability AD.

\section{Conclusion} \label{Conclusion}
{ 

This study introduces CoT-Drive, a novel framework that leverages GPT-4 Turbo with CoT prompt-based engineering to improve scene understanding and prediction accuracy. By utilizing specialized prompt engineering techniques, CoT-Drive enhances the semantic analysis capabilities of LLMs. Additionally, an innovative knowledge distillation strategy allows for the transfer of LLM comprehension to lightweight edge models, enabling them to achieve scene interpretation abilities similar to full LLMs. This approach effectively addresses the computational and latency issues associated with traditional methods, facilitating safe and efficient motion forecasting. To our knowledge, this is the first time a lightweight surrogate model of an LLM has been trained and utilized for motion forecasting tasks. Furthermore, we have developed two new scene description datasets, Highway-Text and Urban-Text, specifically designed to fine-tune lightweight LMs for generating context-specific semantic annotations, thereby optimizing inference costs for real-time operation on edge devices. Comprehensive evaluations on five real-world driving datasets demonstrate that CoT-Drive surpasses state-of-the-art models by significant margins. In summary, CoT-Drive establishes a new benchmark in motion forecasting and provides a practical solution for embedding LLMs in edge AD devices.}

\bibliographystyle{IEEEtran}
\bibliography{IEEEabrv,IEEEexample_new}

\begin{IEEEbiography}
[{\includegraphics[width=1in,height=1.40in, clip,keepaspectratio]{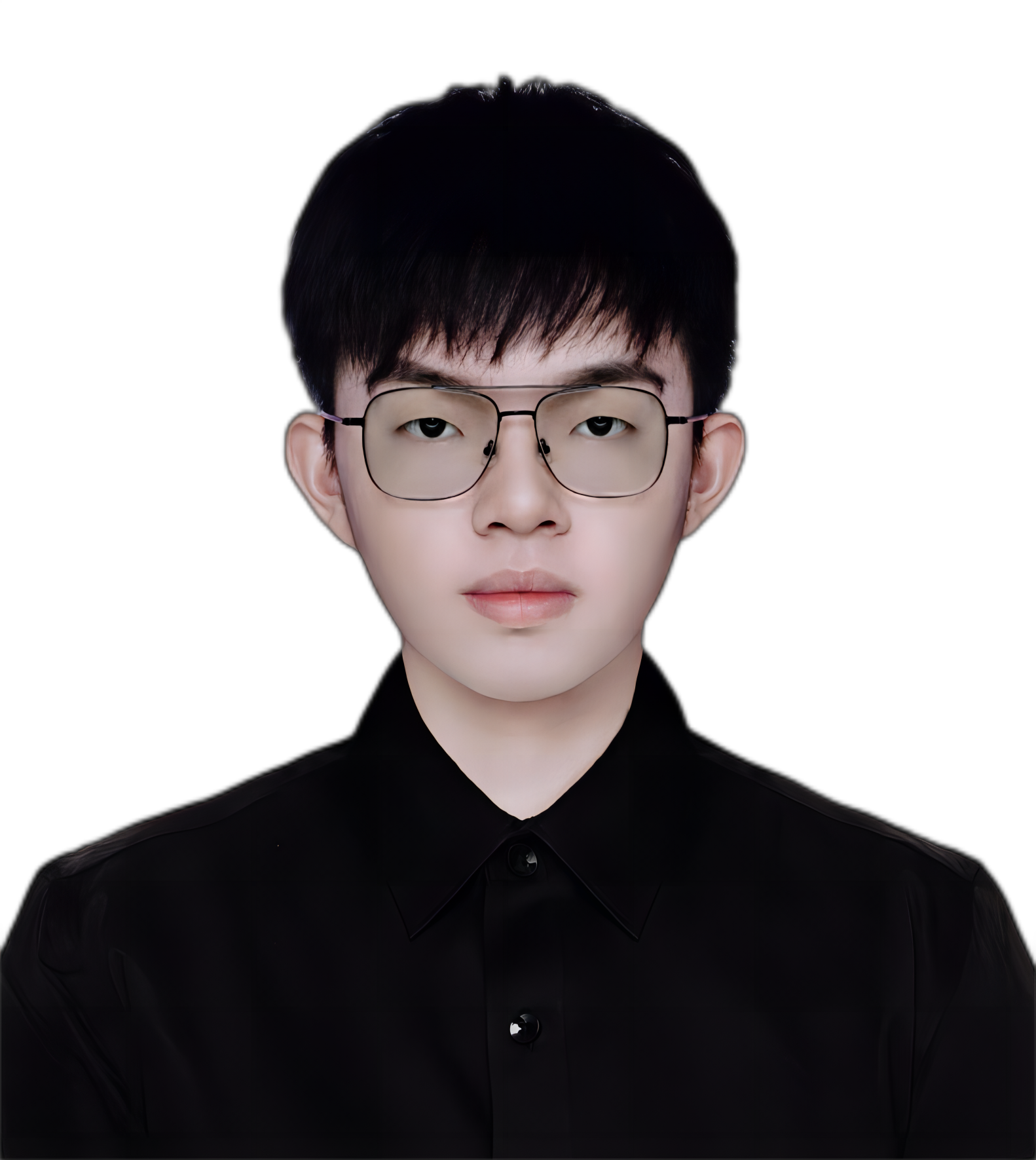}}]{Haicheng Liao} (Student Member, IEEE) received the B.S. degree in software engineering from the University of Electronic Science and Technology of China (UESTC) in 2022. He is currently pursuing a Ph.D. degree at the State Key Laboratory of Internet of Things for Smart City and the Department of Computer and Information Science, University of Macau. Over his academic career, he has published over 20 papers. His research interests include connected autonomous vehicles and the application of deep reinforcement learning to autonomous driving.
\end{IEEEbiography}

\begin{IEEEbiography}
[{\includegraphics[width=1in,height=1.25in, clip,keepaspectratio]{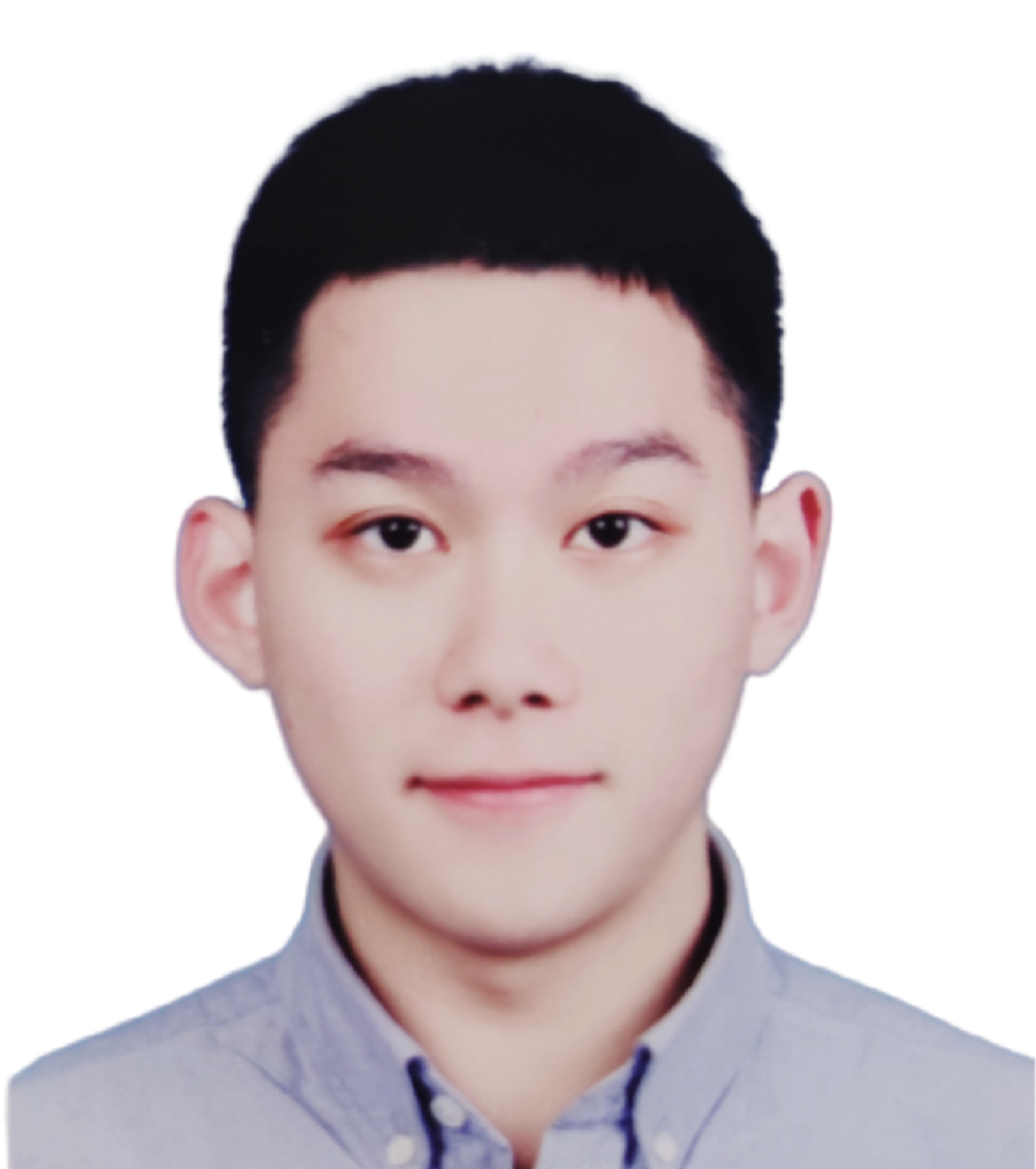}}]  {Hanlin Kong} is an undergraduate student in software engineering from the University of Electronic Science and Technology of China. He is currently serving as a Research Assistant at the State Key Laboratory of Internet of Things for Smart City and the Department of Computer and Information Science, University of Macau. He will pursue a Ph.D. degree at Zhejiang University. His research interests include Driving Planning and trajectory predictions for autonomous driving.
\end{IEEEbiography}

\begin{IEEEbiography}
[{\includegraphics[width=1in,height=1.25in, clip,keepaspectratio]{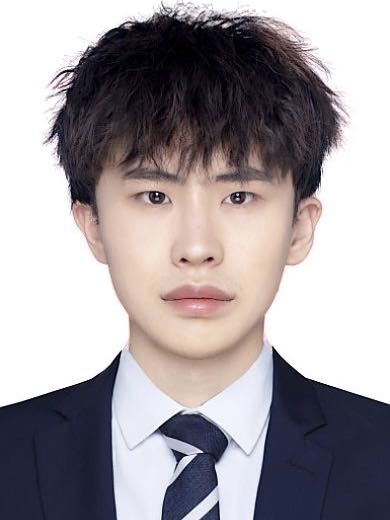}}]{Bonan Wang} Bonan Wang is currently pursuing an M.S. degree at the State Key Laboratory of Internet of Things for Smart City and the Department of Computer and Information Science, University of Macau. He received a B.S. degree in Data Science and Big Data Technology from Shaanxi University of Science \& Technology in 2023. His research interests include trajectory prediction and planning for autonomous driving.
\end{IEEEbiography}

\begin{IEEEbiography}
[{\includegraphics[width=1in,height=1.40in, clip,keepaspectratio]{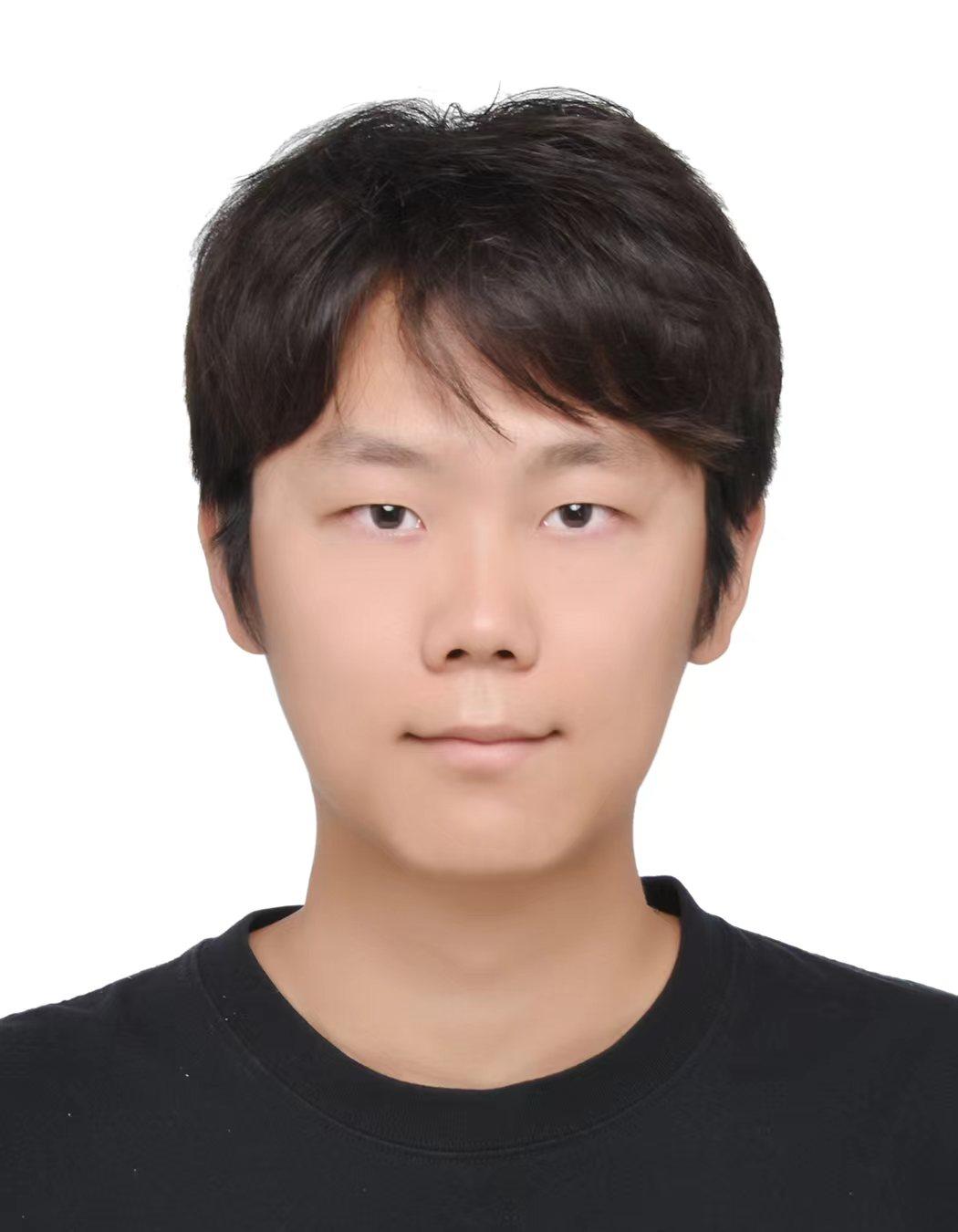}}]{Chengyue Wang} is currently pursuing a Ph.D. degree at the State Key Laboratory of Internet of Things for Smart City and the Department of Civil Engineering, University of Macau. He received his M.S. degree in civil engineering from the University of Illinois Urbana-Champaign (UIUC) in 2022. He received his B.E. degree in transportation engineering from Chang'an University in 2021. His research interests include connected autonomous vehicles and the application of deep reinforcement learning to autonomous driving.
\end{IEEEbiography}

\begin{IEEEbiography}
[{\includegraphics[width=1in,height=1.40in, clip,keepaspectratio]{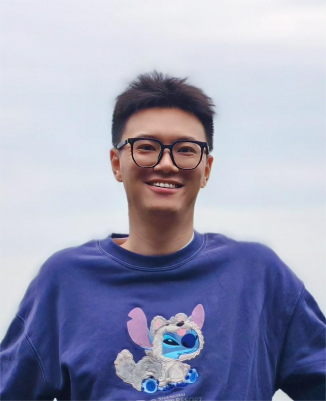}}]{Ye Wang} received a B.S. degree in microelectronics from Peking University, and the M.S. and Ph.D. degree in robotics and the Doctor of Science degree from ETH Zürich.  He is currently an assistant professor at the University of Macau. His research interests include security, blockchain and human-computer interaction.
\end{IEEEbiography}

\begin{IEEEbiography}
[{\includegraphics[width=1in,height=1.25in, clip,keepaspectratio]{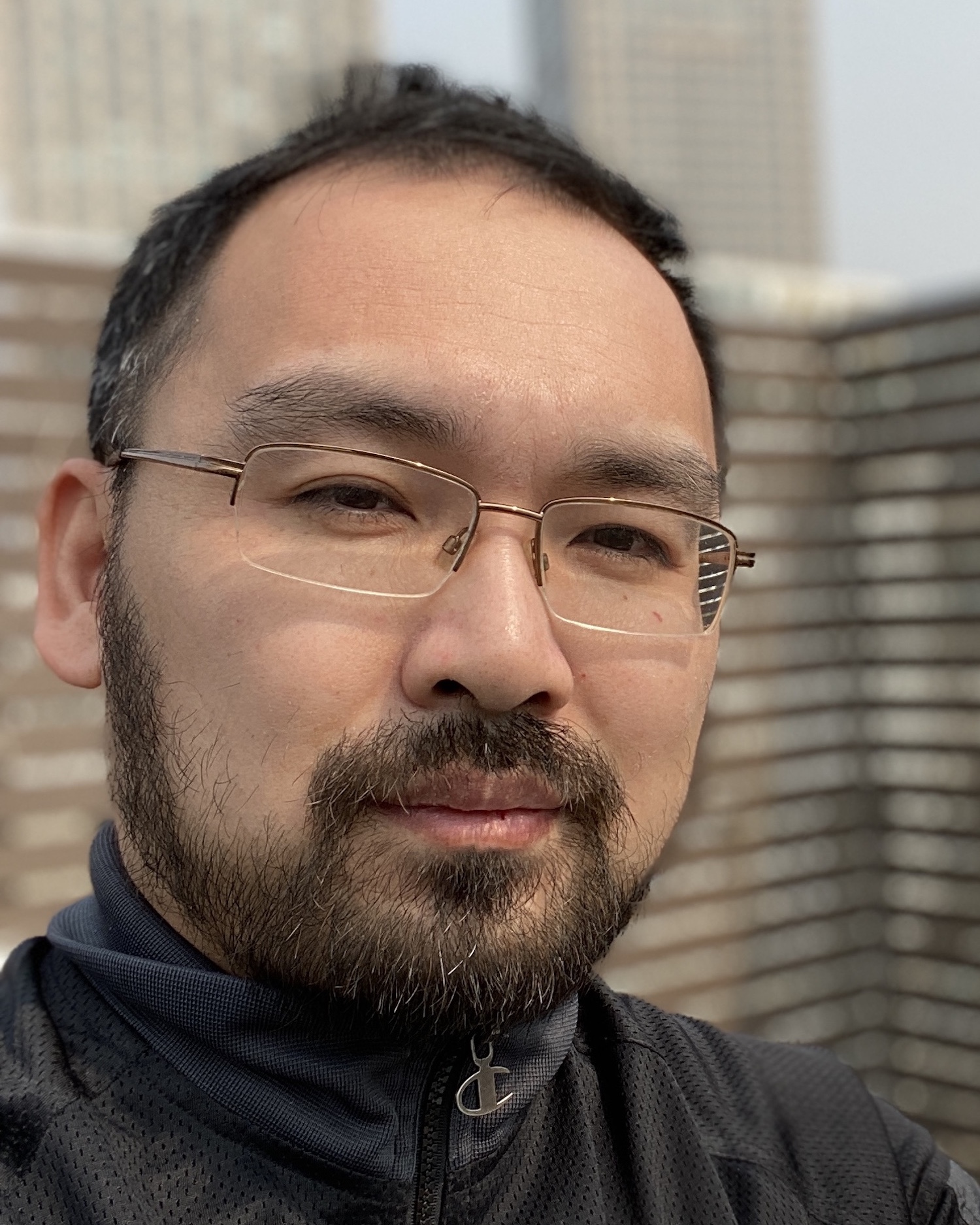}}]{Zhengbing He} (Senior Member, IEEE)  received the Bachelor of Arts degree in English language \& literature from Dalian University of Foreign Languages, China, in 2006, and the Ph.D. degree in Systems Engineering from Tianjin University, China, in 2011. He was a Postdoctoral Researcher and an Assistant Professor at Beijing Jiaotong University, China. From 2018 to 2022, he was a Full Professor at Beijing University of Technology, China. Presently, he is a Senior Research Fellow at Massachusetts Institute of Technology, United States. His research takes traffic flow theory and modeling as a theoretical basis, extending to three cutting-edge branches: (1) data-driven modeling and intelligent vehicle-empowered congestion solution, (2) sensing and prediction of traffic congestion and travel demand, and (3) transportation sustainability-oriented optimization. He has published more than 150 academic papers. He was listed as World’s Top 2\% Scientists. He is the Editor-in-Chief of Journal of Transportation Engineering and Information (Chinese). Meanwhile, he serves as an Associate Editor for IEEE Transactions on Intelligent Transportation Systems, etc., as a Handling Editor for Transportation Research Record, and as an Editorial Advisory Board member for Transportation Research Part C. His webpage is https://www.GoTrafficGo.com and his email is he.zb@hotmail.com.
\end{IEEEbiography}

\begin{IEEEbiography}
[{\includegraphics[width=1in,height=1.25in,clip,keepaspectratio]{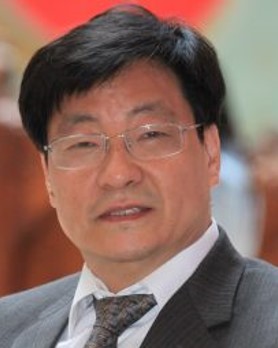}}]{Chengzhong Xu} (Fellow, IEEE) received the Ph.D. degree from The University of Hong Kong, in 1993. He is currently the chair professor of computer science and the dean with the Faculty of Science and Technology, University of Macau. Prior to this, he was with the faculty at Wayne State University, USA, and the Shenzhen Institutes of Advanced Technology, Chinese Academy of Sciences, China. He has published more than 400 papers and more than 100 patents. His research interests include cloud computing and data-driven intelligent applications. He was the Best Paper awardee or the Nominee of ICPP2005, HPCA2013, HPDC2013, Cluster2015, GPC2018, UIC2018, and AIMS2019. He also won the Best Paper award of SoCC2021. He was the Chair of the IEEE Technical Committee on Distributed Processing from 2015 to 2019.
\end{IEEEbiography}

\begin{IEEEbiography}
[{\includegraphics[width=1in,height=1.25in,clip,keepaspectratio]{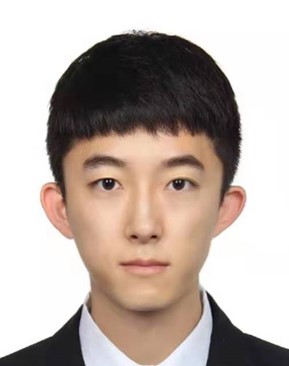}}] {Zhenning Li} (Member, IEEE) received his Ph.D. in Civil Engineering from the University of Hawaii at Manoa, Honolulu, Hawaii, USA, in 2019. Currently, he holds the position of Assistant Professor at the State Key Laboratory of Internet of Things for Smart City, as well as the Department of Computer and Information Science at the University of Macau, Macau.  His main areas of research focus on the intersection of connected autonomous vehicles and Big Data applications in urban transportation systems. He has been honored with several awards, including the Macau Science and Technology Award, Chinese Government Award for Outstanding Self-financed Students Abroad, TRB best young researcher award and the CICTP best paper award, amongst others.
\end{IEEEbiography}

\end{document}